\documentclass[11pt,a4paper]{article}
\usepackage[margin=2cm]{geometry}
\usepackage[T1]{fontenc}
\usepackage[utf8]{inputenc}
\usepackage{stix}

\usepackage{amsmath}
\usepackage{amssymb}
\usepackage[version=4]{mhchem}
\usepackage{siunitx}
\usepackage{booktabs}
\usepackage{array}
\usepackage{longtable}
\usepackage{xcolor}
\usepackage{colortbl}
\usepackage{pifont}
\usepackage{ragged2e}
\usepackage{graphicx}
\usepackage{caption}
\usepackage{placeins}
\usepackage{textcomp}
\usepackage{microtype}
\usepackage{lineno}
\usepackage{authblk}
\usepackage[numbers,sort&compress]{natbib}
\usepackage[hidelinks]{hyperref}

\makeatletter
\newcommand{\prrl@prefix@section}{Supplementary Note }
\newcommand{\prrl@prefix@subsection}{}
\newcommand{\startsupplementary}{%
    \clearpage%
    \setcounter{section}{0}%
    \setcounter{figure}{0}%
    \setcounter{table}{0}%
    \renewcommand{\figurename}{Supplementary Figure}%
    \renewcommand{\tablename}{Supplementary Table}%
    \renewcommand{\theHsection}{si.\arabic{section}}%
    \renewcommand{\theHsubsection}{si.\arabic{section}.\arabic{subsection}}%
    \renewcommand{\theHfigure}{si.\arabic{figure}}%
    \renewcommand{\theHtable}{si.\arabic{table}}%
    \setcounter{secnumdepth}{2}%
    \renewcommand{\@seccntformat}[1]{\@nameuse{prrl@prefix@##1}\@nameuse{the##1}\quad}%
}
\makeatother

\title{Physics-residual reinforcement learning for accurate multi-step temperature prediction of alkaline electrolyzers under renewable power}
\author[1,2]{Yong-Woon Kim}
\author[3,*]{Chulung Kang}
\author[4,*]{Yung-Cheol Byun}
\affil[1]{Department of Computer Engineering, Jeju National University, Jeju, 63243, South Korea}
\affil[2]{Green Hydrogen Global Leading Research Center, Jeju National University, Jeju, 63243, South Korea}
\affil[3]{Department of Mechanical System Engineering, Jeju National University, Jeju, 63243, South Korea}
\affil[4]{Department of Computer Engineering, Major of Electronic Engineering, Food Tech Center (FTC), Jeju National University, Jeju, 63243, South Korea}
\affil[*]{Corresponding authors: Chulung Kang (cukang@jejunu.ac.kr) and Yung-Cheol Byun (ycb@jejunu.ac.kr)}
\date{}

\begin{document}
\maketitle

\begin{abstract}
Alkaline water electrolysis is a key technology for low-emission hydrogen production from renewable electricity, and its variable-power operation requires accurate multi-step electrolyte outlet temperature prediction. Because each predicted value feeds into the next step, a predictor must minimize forecast error and limit the propagation of initial temperature-history differences. We propose physics-residual reinforcement learning (PR-RL), which adds a small reinforcement-learning residual to the state transition of a lumped alkaline electrolysis thermal balance. The residual is trained on the model's own predicted states, so each correction is evaluated by its effect on subsequent steps. A restricted feedback condition preserves the contraction of the physical transition, so initial temperature-history differences decay along the horizon. Evaluated on measured records from two alkaline electrolysis devices against five matched models, PR-RL achieves practical superiority or equivalence in short-term accuracy (mean root-mean-square error at 15 and \SI{30}{\minute} horizons) while reducing the \SI{120}{\minute} residual gain of an initial temperature-history difference by approximately 23-fold to over $10^{8}$-fold relative to reinforcement-learning predictors without the physical transition. These results show that forecast correction and controlled propagation of initial temperature-history differences can be achieved together in one compact model, supporting dynamic operation of renewable-powered alkaline electrolyzers.
\end{abstract}



\section{Introduction}
\label{sec:introduction}

Water electrolysis powered by renewable electricity is a key technology for producing low-emission hydrogen and for linking the power sector with industrial energy demand. The International Energy Agency highlights the growing role of electrolytic hydrogen in expanding low-emission hydrogen supply and in making better use of domestic renewable resources \cite{iea2026}. When an electrolyzer is coupled to wind or solar power, its electric input varies with time, so its performance must be understood not only at the rated operating point but also during load changes \cite{ulleberg2003}. This need places renewed importance on models that describe how the operating state of alkaline water electrolysis (AEL), the most mature electrolysis technology \cite{iea2026}, evolves under dynamic operation, and the stack temperature is a central element of that operating state \cite{ulleberg2003,sakas2022}.

The temperature of an AEL system is governed by electrochemical heat generation, heat exchange with the surroundings, and heat transport through the circulating electrolyte. Because the stack and its auxiliary equipment have considerable thermal capacity and transport delays, the temperature responds to current changes on a much slower time scale than the electrical variables \cite{sakas2022,brauns2023model}, and load changes can produce temperature overshoot \cite{qi2023}. Multi-step temperature prediction therefore has to address two requirements at once. First, the model must represent the heat balance at the current operating point accurately. Second, because each predicted temperature is fed back as an input for the next step, the model must limit how errors accumulate along the prediction horizon \cite{venkatraman2015}. In this recursive setting, the accuracy with respect to the measured temperature and the influence of a difference between the initial temperature histories supplied to two predictions under the same inputs are two distinct properties, and a practically useful predictor should manage both.

Substantial progress has been made on these aspects separately. Physics-based AEL models have evolved from lumped thermal balances \cite{ulleberg2003} to dynamic mass and energy balance models of industrial plants \cite{sakas2022}, validated dynamic models of pressurized systems with electrolyte circulation \cite{brauns2023model}, and higher-order thermal models capturing transport delays for control design \cite{qi2023}. Learning-based prediction has also advanced: a recurrent--Transformer hybrid network predicted temperature for a megawatt-scale AEL plant \cite{wang2026}, and a physics-informed Transformer embedding thermal-balance constraints achieved accurate multi-step prediction for a \SI{5}{\mega\watt} AEL system \cite{ma2026}. For proton exchange membrane (PEM) electrolysis, physics-informed neural networks (PINNs) modeled temperature and identified parameters \cite{zerrougui2025}, and a self-loop PINN demonstrated recursive prediction coupled with model predictive control (MPC) \cite{zerrougui2026}.

A complementary line of research corrects imperfect physical models with learned components. Hybrid process models combining first-principles equations with neural networks are well established \cite{psichogios1992}, and recent frameworks learn corrections on top of known physical dynamics \cite{yin2021}; in thermal applications, grey-box building models were improved by supervised regression of one-step temperature residuals \cite{cilibiu2026}. This physics-residual principle is effective in electrolysis: a neural correction added to deterministic crossover physics predicted the hydrogen crossover of PEM water electrolysis up to pressures beyond the training range \cite{kim2026crossover}, and residual regression on Tafel-based predictions estimated oxygen-evolution catalyst activity from sparse polarization measurements \cite{kim2026catalyst}, in both cases retaining an explicit physical prediction and learning only its correction from measured data. Sequential learning research shows that models trained on one-step targets can behave differently when their own predictions are fed back, and that training on self-predicted states mitigates this mismatch \cite{venkatraman2015}; imposing stability conditions on learned recurrent models has also been studied \cite{revay2020}. Reinforcement learning (RL) has also entered electrolyzer research, but in a different role: RL agents select operating conditions or control commands, such as optimizing the operating point of a zero-gap AEL system via a simulation-based surrogate \cite{lee2026}, or adjusting current commands for PEM electrolyzer fleets via residual RL \cite{nezhadkhatami2026}, following the residual RL concept from robot control \cite{johannink2019}. RL has further been combined with model-based state estimation, where a correction policy updates state estimates from sensor measurements \cite{mowlavi2024}. What has received little attention in this body of work is a prediction-oriented configuration in which the state transition of a simple AEL thermal balance is retained, and the temperature residual added to that transition is learned from a reward formed from the subsequent prediction errors evaluated on the model's own predicted states. How much of the resulting behavior comes from the physical transition, how much from the sequential learning scheme, and how much from their combination---in particular with respect to the joint handling of prediction accuracy and the damping of initial temperature-history differences---has likewise received little attention under matched conditions.

To address this gap, we propose physics-residual reinforcement learning (PR-RL), a predictor that adds a single small reinforcement-learning residual policy to the state transition of a lumped AEL thermal balance. The residual policy is trained on the model's own predicted states using a reward formed from the subsequent temperature prediction errors, so that each correction is evaluated by its effect on the following steps rather than by a one-step fit alone. The feedback of the residual to the temperature history is deliberately restricted, which preserves the damping condition that the physical transition provides for differences in the initial temperature history.

We evaluate PR-RL on measured operating records of two different AEL devices, a pressurized single-cell system \cite{brauns2023data} and a three-cell short stack \cite{peace2026}, against five matched comparison models that share the same inputs and test conditions and, where applicable, the same physical parameters. Compared with the physics-based model and with a conventional supervised residual multilayer perceptron (MLP), PR-RL achieves superiority or practical equivalence in a short-term error score, defined as the mean of the root mean square error (RMSE) values at the 15 and \SI{30}{\minute} horizons, and it reduces the 95th percentile of the trajectory maximum absolute error (P95) over \SI{30}{\minute} horizons relative to the physics-based model on both datasets. With the same trained weights, PR-RL reduces the \SI{120}{\minute} persistence of an initial temperature-history difference compared with two RL predictors that have no physical transition, while remaining non-inferior in the short-term error score relative to both within an engineering margin of \SI{0.10}{\degreeCelsius} adopted in this study.

This study makes three contributions. The first is the contribution of the physical transition: we combine the AEL thermal-balance transition with a restricted residual feedback and derive a sufficient condition under which differences in the initial temperature history decay along the prediction horizon. Evaluation on the measured records confirms that, compared with an RL predictor trained with the same learning scheme without the physical transition, both the maximum propagation and the \SI{120}{\minute} persistence of such differences are reduced. The second is the contribution of the reinforcement-learned residual: we correct the prediction error of the same physical model through sequentially rewarded residual learning and quantify the effect of this learning scheme relative to a directly supervised residual MLP that uses the identical physical model, inputs, network architecture, and damping constraint, with a Random Forest residual model \cite{breiman2001} as an additional reference. The third is the contribution of the combination: we show that a single model consisting of the physical transition and one residual policy network jointly provides non-inferior short-term accuracy relative to both RL predictors without the physical transition and a lower \SI{120}{\minute} persistence of initial temperature-history differences.

By uniting the damping property of a simple physical transition with reward-based residual correction, this study provides a design principle for AEL temperature predictors that delivers accurate multi-step prediction and controlled propagation of initial temperature-history differences within one compact model, supporting the reliable dynamic operation of renewable-powered alkaline electrolyzers.

\section{Results}
\label{sec:results}

\subsection{PR-RL framework and evaluation design}
\label{sec:res-overview}

\begin{figure}[!tp]
\def\baselinestretch{1}\normalsize
\centering
\includegraphics[width=0.99\textwidth]{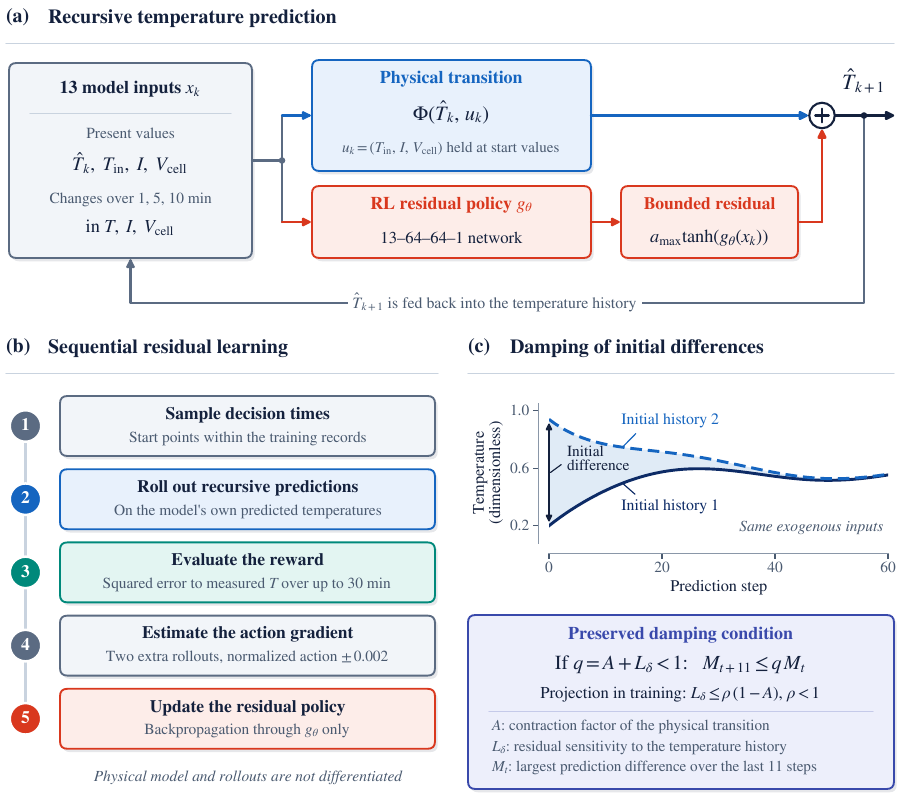}
\caption{\textbf{Overview of the PR-RL temperature prediction framework.}
(a)~Inference structure (equation~\eqref{eq:prrl}; inputs defined in the `Prediction task and datasets' section of Methods).
(b)~Training signal flow (the `PR-RL: reinforcement-learning residual policy' section of Methods); numbered steps give the order of the training operations. In~(a) and~(b), the colored boxes distinguish the learned residual policy (red), the physical transition (blue), and the measured-temperature reward signal (green).
(c)~Damping of initial temperature-history differences under the condition $q<1$ (the `Damping of initial temperature-history differences' section of Methods), illustrated with example trajectories computed from the discrete transition for generic parameter values on dimensionless axes; shading marks the pointwise difference between the two predictions.}
\label{fig:framework}
\end{figure}

The prediction task is the recursive multi-step prediction of a representative electrolyte outlet temperature of an AEL system at \SI{60}{\second} steps: each predicted temperature is fed back as an input for the next step, while the exogenous operating inputs are held at their values at the prediction start time. Two public measured datasets are used: the Brauns dataset, published by Brauns and Turek for a pressurized single-cell alkaline electrolysis system, and the PEACE dataset, published by the Pressurized Efficient Alkaline EleCtrolysEr (PEACE) project for a pressurized three-cell short stack. The PR-RL model combines a lumped thermal-balance transition, identified from training data, with one bounded residual produced by a reinforcement-learning residual policy network (Fig.~\ref{fig:framework}).

The residual policy is trained as a deterministic policy on the model's own predicted states: the model rolls out predictions from a decision time, and the action value of a residual action is estimated from the squared errors between the subsequent predicted and measured temperatures. The physical transition contracts temperature differences by a per-step factor $A<1$; after every optimizer update a weight projection limits the sensitivity $L_\delta$ of the residual to the temperature history, so that $q = A + L_\delta$ remains below one. Under this condition, a difference between the initial temperature histories of two predictions driven by the same exogenous inputs decays along the prediction horizon.

Five comparison models are evaluated under matched conditions: the physics-based model (the physical transition alone), the physics-based model with supervised residual MLP (a residual network trained by direct supervised regression of the measured one-step residual), the physics-based model with Random Forest residual (the same one-step residual target learned by a Random Forest), the matched RL-only model (the same deterministic policy learning applied to the full temperature change without the physical transition), and the TD3-only model (a predictor without the physical transition trained with twin delayed deep deterministic policy gradient (TD3)). Prediction accuracy is assessed by the short-term error score $S$ (the mean of the root mean square errors at the 15 and \SI{30}{\minute} horizons) and by P95 (the 95th percentile over start points of the maximum absolute error within each \SI{30}{\minute} trajectory). The propagation of initial temperature-history differences is evaluated by adding \SI{0.5}{\degreeCelsius} to the initial temperature history of one of two otherwise identical predictions; the gain is the absolute difference between the two predicted trajectories divided by this initial difference. The maximum propagation gain is the 95th percentile, over start points, of the maximum gain within each \SI{120}{\minute} trajectory, and the \SI{120}{\minute} residual gain is the 95th percentile, over start points, of the gain remaining at \SI{120}{\minute}. For the error metrics $S$ and P95, practical superiority, equivalence, and inferiority are judged with a margin of $\delta = \SI{0.10}{\degreeCelsius}$. Details are in Methods.

All comparisons in this section follow the decision rules of the `Statistical analysis' section of Methods. Differences are defined as the PR-RL value minus the comparison-model value; smaller values are better for every metric. The reported intervals are the Bonferroni-adjusted simultaneous confidence intervals (CIs) for each comparison family. Each comparison is based on the twenty training seeds of the main cohort, except for the PEACE comparison against the supervised residual MLP, which uses its separate confirmation cohort of 400 seeds.

\subsection{Damping of initial temperature-history differences}
\label{sec:res-damping}

We first evaluate the contribution of the physics-based structure by comparing PR-RL with the matched RL-only model, which uses the same learning scheme without the physical transition, and additionally with the TD3-only model (Methods).

Table~\ref{tab:gains} reports the propagation of an initial temperature-history difference of \SI{0.5}{\degreeCelsius} under identical exogenous inputs.

\begin{table}[!htbp]
\def\baselinestretch{1}\normalsize\small
\caption{\textbf{Propagation and persistence of an initial temperature-history difference.} Propagation and persistence of an initial temperature-history difference of \SI{0.5}{\degreeCelsius}: PR-RL versus the matched RL-only model and the TD3-only model. Each entry is the dimensionless 95th-percentile gain over start points (the `Evaluation metrics' section of Methods), averaged with equal weights over the held-out sessions within each seed and then over the twenty training seeds. The difference is defined as PR-RL minus the comparison model, and the interval is the Bonferroni-adjusted simultaneous confidence interval of the gain family (eight comparisons, error probability 5\%). Superiority (inferiority) requires the simultaneous upper (lower) limit to be below (above) zero; no temperature margin is applied to the dimensionless gains.}
\label{tab:gains}
\centering
\small
\resizebox{\textwidth}{!}{%
\begin{tabular}{@{}lllccccc@{}}
\toprule
Dataset & Comparison model & Metric & PR-RL & Comparison & Difference & Simultaneous CI & Judgment \\
\midrule
Brauns & Matched RL-only model & Maximum propagation gain & $0.9425$ & $2.2496$ & $-1.3071$ & $[-1.4827,\,-1.1315]$ & Superior \\
Brauns & Matched RL-only model & \SI{120}{\minute} residual gain & $0.0074$ & $1.1785$ & $-1.1711$ & $[-1.3502,\,-0.9920]$ & Superior \\
Brauns & TD3-only model & Maximum propagation gain & $0.9425$ & $0.7618$ & $+0.1807$ & $[+0.1420,\,+0.2193]$ & Inferior \\
Brauns & TD3-only model & \SI{120}{\minute} residual gain & $0.0074$ & $0.1680$ & $-0.1607$ & $[-0.2520,\,-0.0694]$ & Superior \\
PEACE & Matched RL-only model & Maximum propagation gain & $0.7870$ & $1.0457$ & $-0.2587$ & $[-0.2887,\,-0.2286]$ & Superior \\
PEACE & Matched RL-only model & \SI{120}{\minute} residual gain & $4.2\times10^{-9}$ & $0.8200$ & $-0.8200$ & $[-0.9169,\,-0.7231]$ & Superior \\
PEACE & TD3-only model & Maximum propagation gain & $0.7870$ & $0.9506$ & $-0.1636$ & $[-0.2707,\,-0.0565]$ & Superior \\
PEACE & TD3-only model & \SI{120}{\minute} residual gain & $4.2\times10^{-9}$ & $0.6244$ & $-0.6244$ & $[-0.8388,\,-0.4101]$ & Superior \\
\bottomrule
\end{tabular}}
\end{table}

 On both datasets, PR-RL markedly reduces the maximum propagation gain and the \SI{120}{\minute} residual gain relative to the matched RL-only model. The maximum propagation gain decreases from 2.2496 to 0.9425 on the Brauns dataset and from 1.0457 to 0.7870 on the PEACE dataset: the matched RL-only gain exceeds one, indicating amplification beyond the initial difference, whereas the PR-RL gain stays below one on both datasets, indicating attenuation. The \SI{120}{\minute} residual gain decreases from 1.1785 to 0.0074 and from 0.8200 to $4.2\times10^{-9}$, respectively, so two hours after the prediction start the initial difference has essentially vanished from the PR-RL predictions, whereas a large share of it persists in the matched RL-only predictions. Relative to the TD3-only model, the \SI{120}{\minute} residual gain of PR-RL is also lower on both datasets, and the maximum propagation gain is lower on PEACE; on Brauns, the TD3-only model attains the smaller maximum propagation gain, with both values below one (Table~\ref{tab:gains}). Seven of the eight simultaneous intervals establish statistical superiority for PR-RL.

These observed reductions reflect the structural damping condition established in Methods. Every one of the 140 PR-RL models of the main cohort (seven leave-one-session-out conditions times twenty seeds) satisfies the damping bound $q<1$, with a maximum of 0.98684, and the identically constrained supervised residual MLP satisfies the same condition in all of its 140 models, with a maximum of 0.98359. The per-condition parameters, projection limits, and complete $q$ distributions are given in Supplementary Note~\ref{app:physparams} (Supplementary Fig.~\ref{fig:damping}). The damping is therefore a property of the physics-based structure: the physical transition together with the bounded and projected residual acting on it. This property is preserved whether the residual is trained by reinforcement learning or by supervised regression. The 140 values for each learning method correspond to trained models under the fixed operating records, not independent devices.

\subsection{Temperature prediction accuracy and the effect of residual learning}
\label{sec:res-accuracy}

The two datasets differ in their target-temperature ranges and variability. On the common \SI{60}{\second} grid, the target temperature spans 58.11--\SI{59.61}{\degreeCelsius} in the Brauns records and 14.94--\SI{32.71}{\degreeCelsius} in the PEACE records (Methods), with overall standard deviations of about \SI{0.30}{\degreeCelsius} and \SI{3.59}{\degreeCelsius}, respectively.

\begin{figure}[!htbp]
\def\baselinestretch{1}\normalsize
\centering
\includegraphics[width=0.92\textwidth]{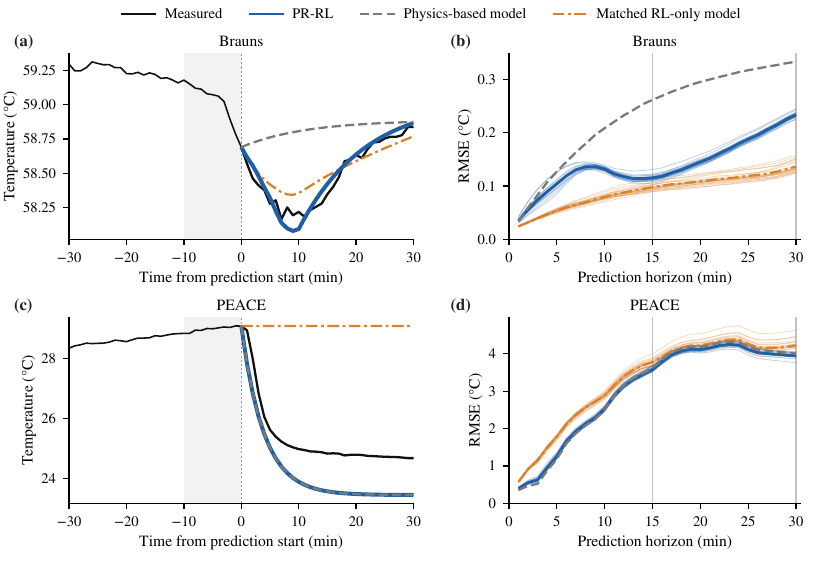}
\caption{\textbf{Measured and predicted electrolyte outlet temperatures.} Measured and predicted electrolyte outlet temperatures for the Brauns (a,~b) and PEACE (c,~d) datasets. (a,~c)~A representative \SI{30}{\minute} forecast of PR-RL, the physics-based model, and the matched RL-only model in the held-out session, selected by a fixed median rule (the seed whose PR-RL score $S$ is closest to the median over the twenty seeds of the held-out session, and within that seed, the start point whose \SI{30}{\minute} trajectory RMSE is closest to the median over all start points); the shaded interval marks the \SI{10}{\minute} measured temperature history used as model input. In~(c), the PR-RL and physics-based curves coincide because the selected checkpoint has zero residual output (see text). (b,~d)~RMSE at each prediction horizon, averaged over the held-out sessions and the twenty training seeds (thick curves); thin curves are the session-averaged results of individual seeds, not confidence bands. The vertical lines mark the 15 and \SI{30}{\minute} horizons whose RMSE average defines $S$. Selected examples: Brauns, record of 2021-07-05, seed 104, start 2021-07-06 08:03 Coordinated Universal Time (UTC), one of 759 start points; PEACE, experiment 3, seed 101, start 2026-02-04 14:00 UTC, one of 41 start points. The horizon-wise RMSE is averaged with equal weights over the held-out sessions within each seed and then over the twenty training seeds; the physics-based model has no seed dependence.}
\label{fig:prediction}
\end{figure}

Fig.~\ref{fig:prediction} shows a representative \SI{30}{\minute} forecast and the horizon-wise RMSE of PR-RL, the physics-based model, and the matched RL-only model on each dataset. The forecast examples are selected by a fixed median rule (legend of Fig.~\ref{fig:prediction}). In the PEACE example, the PR-RL and physics-based predictions coincide because the checkpoint selected for this seed is the initial network with zero residual output, as in 18 of the 20 seeds of that session (Supplementary Fig.~\ref{fig:si-checkpoints}).

Table~\ref{tab:accuracy} reports all twenty final accuracy comparisons.

\begin{table}[!htbp]
\def\baselinestretch{1}\normalsize\footnotesize
\caption{\textbf{Final accuracy comparisons of PR-RL.} Final accuracy comparisons of PR-RL with the five comparison models: (a)~short-term error score $S$ and (b)~trajectory maximum error P95, both in \si{\degreeCelsius}. Values are averaged with equal weights over the held-out sessions within each seed and then over the training seeds. Differences are defined as PR-RL minus the comparison model, with Bonferroni-adjusted simultaneous confidence intervals (the ten comparisons of each metric form one family with error probability 2.5\%; the PEACE comparison against the supervised residual MLP forms a separate confirmation family of two comparisons at 2.5\%). Practical judgments use the \SI{0.10}{\degreeCelsius} margin of the `Statistical analysis' section of Methods; the statistical direction is superior when the simultaneous upper limit is below zero, inferior when the lower limit is above zero, and undecided when the interval contains zero. The PEACE rows against the supervised residual MLP report the separate 400-seed confirmation cohort (the `Statistical analysis' section of Methods). PR-RL is non-inferior within the \SI{0.10}{\degreeCelsius} margin in every comparison except the Brauns P95 comparison with the matched RL-only model and the PEACE P95 comparison with the TD3-only model.}
\label{tab:accuracy}
\centering
\footnotesize
\setlength{\tabcolsep}{4pt}
\resizebox{\textwidth}{!}{%
\begin{tabular}{@{}llccccccc@{}}
\toprule
Dataset & Comparison model & Seeds & PR-RL & Comparison & Difference & Simultaneous CI & Practical & Statistical \\
\midrule
\multicolumn{9}{@{}l}{(a) \itshape Short-term error score $S$ (\si{\degreeCelsius})} \\
Brauns & Physics-based model & 20 & 0.1746 & 0.2976 & $-0.1229$ & $[-0.1269,\,-0.1190]$ & Superior & Superior \\
Brauns & Physics + supervised residual MLP & 20 & 0.1746 & 0.2980 & $-0.1233$ & $[-0.1278,\,-0.1188]$ & Superior & Superior \\
Brauns & Physics + Random Forest residual & 20 & 0.1746 & 0.2999 & $-0.1252$ & $[-0.1291,\,-0.1214]$ & Superior & Superior \\
Brauns & Matched RL-only model & 20 & 0.1746 & 0.1174 & $+0.0573$ & $[+0.0504,\,+0.0642]$ & Equivalent & Inferior \\
Brauns & TD3-only model & 20 & 0.1746 & 0.2835 & $-0.1089$ & $[-0.1157,\,-0.1021]$ & Superior & Superior \\
PEACE & Physics-based model & 20 & 3.7616 & 3.8122 & $-0.0506$ & $[-0.0938,\,-0.0074]$ & Equivalent & Superior \\
PEACE & Physics + supervised residual MLP & 400 & 3.7572 & 3.7040 & $+0.0532$ & $[+0.0387,\,+0.0677]$ & Equivalent & Inferior \\
PEACE & Physics + Random Forest residual & 20 & 3.7616 & 3.9540 & $-0.1924$ & $[-0.2357,\,-0.1490]$ & Superior & Superior \\
PEACE & Matched RL-only model & 20 & 3.7616 & 3.9947 & $-0.2331$ & $[-0.3317,\,-0.1345]$ & Superior & Superior \\
PEACE & TD3-only model & 20 & 3.7616 & 4.0875 & $-0.3260$ & $[-0.5505,\,-0.1014]$ & Superior & Superior \\
\addlinespace
\multicolumn{9}{@{}l}{(b) \itshape Trajectory maximum error P95 (\si{\degreeCelsius})} \\
Brauns & Physics-based model & 20 & 0.5767 & 0.8578 & $-0.2811$ & $[-0.2916,\,-0.2706]$ & Superior & Superior \\
Brauns & Physics + supervised residual MLP & 20 & 0.5767 & 0.8759 & $-0.2992$ & $[-0.3116,\,-0.2868]$ & Superior & Superior \\
Brauns & Physics + Random Forest residual & 20 & 0.5767 & 0.9985 & $-0.4218$ & $[-0.4328,\,-0.4107]$ & Superior & Superior \\
Brauns & Matched RL-only model & 20 & 0.5767 & 0.3335 & $+0.2432$ & $[+0.2240,\,+0.2624]$ & Inferior & Inferior \\
Brauns & TD3-only model & 20 & 0.5767 & 0.8588 & $-0.2821$ & $[-0.3179,\,-0.2462]$ & Superior & Superior \\
PEACE & Physics-based model & 20 & 8.4861 & 8.7867 & $-0.3006$ & $[-0.4102,\,-0.1911]$ & Superior & Superior \\
PEACE & Physics + supervised residual MLP & 400 & 8.4800 & 8.4460 & $+0.0339$ & $[-0.0027,\,+0.0706]$ & Equivalent & Undecided \\
PEACE & Physics + Random Forest residual & 20 & 8.4861 & 8.8121 & $-0.3260$ & $[-0.4316,\,-0.2204]$ & Superior & Superior \\
PEACE & Matched RL-only model & 20 & 8.4861 & 8.9301 & $-0.4440$ & $[-0.7173,\,-0.1706]$ & Superior & Superior \\
PEACE & TD3-only model & 20 & 8.4861 & 8.5420 & $-0.0559$ & $[-0.7014,\,+0.5895]$ & Undecided & Undecided \\
\bottomrule
\end{tabular}}
\end{table}

Adding the reinforcement-learning residual to the same physical model improves both error metrics on the Brauns dataset: $S$ decreases from 0.2976 to \SI{0.1746}{\degreeCelsius} (difference $-0.1229$; a 41.3\% reduction) and P95 from 0.8578 to \SI{0.5767}{\degreeCelsius} ($-0.2811$; a 32.8\% reduction), both practically superior. On the PEACE dataset, $S$ decreases by \SI{0.0506}{\degreeCelsius} (1.3\%), a statistically superior and practically equivalent result, and P95 decreases by \SI{0.3006}{\degreeCelsius} (3.4\%), a practically superior result. The consistent reduction in P95 on both devices demonstrates effective correction of large prediction errors by the learned residual.

The comparison with the supervised residual MLP isolates the learning scheme, since both models share the physical parameters, inputs, network architecture, and damping projection (Methods). On the Brauns dataset, PR-RL is practically superior in both metrics ($S$: $-0.1233$; P95: $-0.2992$). On the PEACE dataset, the confirmation cohort of 400 seeds establishes practical equivalence in both metrics, with mean differences below \SI{0.06}{\degreeCelsius} (a small statistical inferiority in $S$ and an undecided statistical direction in P95) on a dataset whose target temperature spans almost \SI{18}{\degreeCelsius}. Relative to direct one-step residual regression, sequentially rewarded learning on self-predicted states thus provides a clear error reduction on Brauns while fully maintaining practically equivalent accuracy on PEACE.

The frequent selection of the initial checkpoint explains why the supervised MLP errors in Table~\ref{tab:accuracy} are close to those of the physics-based model. Because the last network layer is initialized to zero, the epoch-0 checkpoint has a zero residual output and reproduces the physics-based model exactly. This initial checkpoint is retained whenever no updated candidate improves the validation score $S$. The initial checkpoint was retained for the supervised MLP in 57 of 80 Brauns runs (71.25\%) and for PR-RL in 0 of 80; see Supplementary Table~\ref{tab:si-epoch0} and Supplementary Fig.~\ref{fig:si-checkpoints} for all cohorts.

PR-RL is also practically superior to the Random Forest residual in all four comparisons ($S$ and P95 on both datasets); because that contrast additionally changes the function class, the matched MLP comparison provides the most direct evidence on the learning scheme. Against the TD3-only model, PR-RL is practically superior in $S$ on both datasets and in P95 on Brauns; the PEACE P95 comparison is undecided at the \SI{0.10}{\degreeCelsius} margin because of its wide simultaneous interval (Table~\ref{tab:accuracy}). Against the matched RL-only model, PR-RL is practically superior in both metrics on the PEACE dataset. On the Brauns dataset, the matched RL-only model attains a lower $S$ (0.1174 versus \SI{0.1746}{\degreeCelsius}, within the practical-equivalence margin) and a lower P95 (0.3335 versus \SI{0.5767}{\degreeCelsius}, a practical inferiority of PR-RL). The joint comparison below shows that, with the same trained weights, PR-RL combines this practically equivalent $S$ with substantially stronger damping. Session-level details of all comparisons are provided in Supplementary Fig.~\ref{fig:si-sessions}, and the sensitivity of the practical judgments to the margin is shown in Supplementary Fig.~\ref{fig:si-margins}.

\subsection{Joint prediction accuracy and damping}
\label{sec:res-joint}

The third contribution concerns whether the same trained PR-RL weights provide both the accuracy and the damping described above.

\begin{figure}[!htbp]
\def\baselinestretch{1}\normalsize
\centering
\includegraphics[width=0.9\textwidth]{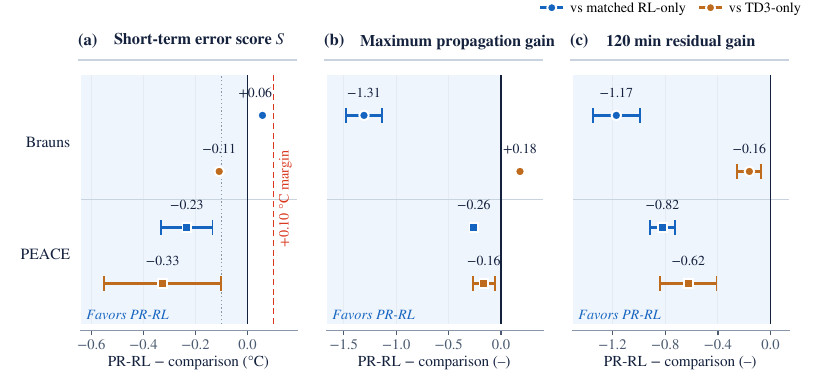}
\caption{\textbf{Paired differences between PR-RL and the two RL-only models.} Paired differences between PR-RL and the matched RL-only model (blue) and between PR-RL and the TD3-only model (orange) in (a)~the short-term error score $S$, (b)~the maximum propagation gain, and (c)~the \SI{120}{\minute} residual gain, for the Brauns and PEACE datasets. Differences are PR-RL minus the comparison model, so negative values (shaded region) favor PR-RL. Markers are mean paired differences and numbers give their values; error bars are the family-adjusted simultaneous confidence intervals of the `Statistical analysis' section of Methods, some narrower than the marker. In~(a), the dashed line marks the $+\SI{0.10}{\degreeCelsius}$ non-inferiority margin, the dotted line $-\SI{0.10}{\degreeCelsius}$, and the solid line zero.}
\label{fig:joint}
\end{figure}

Fig.~\ref{fig:joint} summarizes the paired differences between PR-RL and the two predictors without the physical transition in $S$, the maximum propagation gain, and the \SI{120}{\minute} residual gain, evaluated with the same fitted model in each session--seed condition of the main cohort (twenty training seeds), so that the accuracy and both gain metrics refer to identical trained weights. The gain differences are dimensionless and concern the propagation of initial temperature-history differences between two predictions under the same inputs, not the error with respect to the measured temperature.

 On both datasets, the simultaneous confidence intervals of the $S$ differences lie below the \SI{0.10}{\degreeCelsius} non-inferiority margin for both comparison models (matched RL-only: Brauns $[+0.0504, +0.0642]$, PEACE $[-0.3317, -0.1345]$; TD3-only: Brauns $[-0.1157, -0.1021]$, PEACE $[-0.5505, -0.1014]$), and the simultaneous upper limits of the \SI{120}{\minute} residual-gain differences are below zero in all four dataset--comparison pairs. The same PR-RL models therefore deliver non-inferior $S$ relative to both predictors without the physical transition together with a decisive reduction of the \SI{120}{\minute} persistence of initial temperature-history differences on both datasets; relative to the matched RL-only model, the maximum propagation gain is reduced as well.

The Brauns dataset demonstrates the benefit of combining accuracy and damping: the matched RL-only model attains a mean $S$ advantage of about \SI{0.0573}{\degreeCelsius}, which stays within the practical-equivalence margin, while PR-RL lowers the average \SI{120}{\minute} residual gain from 1.1785 to 0.0074, a reduction in the persistence of initial differences by two orders of magnitude; relative to the TD3-only model the same reduction is about 23-fold, although that model attains the smaller maximum propagation gain on this dataset (Table~\ref{tab:gains}), which is why short-term accuracy, the maximum propagation, and the long-term persistence are assessed together. The non-inferiority refers to the score $S$ at the \SI{0.10}{\degreeCelsius} margin (Methods); Table~\ref{tab:accuracy} reports the P95 comparisons separately. Together with the residual-learning results above, these findings demonstrate the three contributions: the physical transition with restricted residual feedback provides the damping, the sequentially rewarded residual provides the error correction, and their combination delivers both properties within one compact model.

Supplementary Notes~\ref{si:extrap}--\ref{si:long} characterize how prediction accuracy varies across operating domains, devices, and prediction horizons beyond those used for training and selection, and Supplementary Note~\ref{si:stats} provides the session-level, cohort, and margin-sensitivity details that support the results of this section.

\section{Discussion}
\label{sec:discussion}

The damping results follow from two structural elements: the contraction of the thermal transition and the restriction of the residual feedback. In the lumped heat balance, positive heat capacity and heat transfer coefficients give a transition factor $A<1$, which contracts temperature differences by a fixed factor at every step under the same exogenous inputs. A learned residual can weaken this contraction, because its correction depends on the predicted temperature history and therefore responds to any difference between two histories. The projection restricts this feedback so that the combined bound $q=A+L_\delta$ remains below one. All 140 PR-RL models of the main cohort satisfy this condition, and so do all 140 identically constrained supervised residual MLP models (Supplementary Note~\ref{app:physparams}). Because both learning methods satisfy the same condition, the damping is attributable to the structure, namely the physical transition with restricted residual feedback, rather than to the learning scheme. Two restrictions act on the residual and should be distinguished: the output bound $a_{\max}$ limits only the size of each correction, whereas the projection limits the sensitivity $L_\delta$ of the correction to the predicted temperature history, and it is this restriction that secures $q<1$. This design adapts the contraction conditions developed for learned recurrent models \cite{revay2020} to a physics-based hybrid predictor, where the identified heat balance itself supplies the contraction and the projection only preserves it. The TD3-only model, which is not covered by the condition, also attenuates the evaluated differences and attains a lower maximum propagation gain on Brauns, whereas for PR-RL the decay is an explicit, verifiable property rather than an outcome checked only empirically after training.

The accuracy results demonstrate that a learned residual can correct the prediction errors of the lumped heat balance without replacing it. The physical model represents electrochemical heat generation, heat loss, and heat exchange with the incoming electrolyte, while the residual adjusts the next predicted temperature. This division of roles follows the hybrid modeling approach of augmenting physical dynamics with a learned correction \cite{psichogios1992,yin2021}. PR-RL extends the supervised physics-residual designs demonstrated for electrolysis \cite{kim2026crossover,kim2026catalyst} to recursive multi-step temperature prediction. The methodological contribution is to couple sequential reward learning with a bound on the residual's sensitivity to the temperature history, set from the contraction factor of the physical transition, so that residual learning preserves an explicit damping condition. Because PR-RL and the physics-based model use identical identified parameters, their comparison directly quantifies the improvement from the learned correction. The P95 reduction on both datasets shows that the residual corrects the largest forecast errors of the physical model. The Brauns results additionally show a substantial reduction in $S$, and the PEACE results show a statistically superior $S$ within the practical-equivalence margin together with a practically superior P95 (Table~\ref{tab:accuracy}). On PEACE the change in $S$ is small while P95 improves, so the two metrics reveal complementary aspects of the correction.

Sequential reward learning evaluates each correction by its effect on later prediction steps. A correction changes the predicted temperature fed back into the model and therefore changes the states on which subsequent corrections act. The supervised residual MLP fits one-step residuals at measured states, whereas PR-RL learns from rewards evaluated along its own recursive predictions. This distinction addresses the known mismatch between one-step fitting and recursive prediction \cite{venkatraman2015}. The matched RL-only model separates the contribution of the physics-based structure under the same learning procedure, and the TD3-only model extends this comparison to a representative actor--critic algorithm without the physical transition. With physical parameters, inputs, network architecture, and damping projection matched, the MLP comparison shows practical superiority of PR-RL on Brauns and practical equivalence on PEACE in both $S$ and P95; this comparison isolates the learning strategy, and the advantage on Brauns demonstrates the benefit of learning corrections from sequential rewards rather than from direct one-step regression. Both methods also include a built-in safeguard: the validation-based selection can fall back on the initial zero-residual checkpoint, so a learned correction is deployed only when it improves the validation score (Supplementary Fig.~\ref{fig:si-checkpoints}). The Random Forest results extend the comparison to a second direct residual regression method. Together, these comparisons support the central design choice of PR-RL: corrections are learned from their effects on subsequent forecasts while the physical damping condition is preserved.

Prediction accuracy and the damping of initial temperature-history differences are complementary properties of a recursive predictor. Evaluating both properties with the same fitted weights in each condition and seed shows that error correction and damping can be achieved together in one PR-RL model. On both datasets, the paired comparisons establish non-inferiority in $S$ relative to the matched RL-only model and superiority in both the maximum propagation gain and the \SI{120}{\minute} residual gain (Fig.~\ref{fig:joint}). On Brauns, the difference in $S$ stays within the \SI{0.10}{\degreeCelsius} practical margin, while PR-RL reduces the average \SI{120}{\minute} residual gain by a factor of about 160. This combination matters in operation: the measured temperature history supplied at the start of a forecast can contain sensor noise or data gaps, so two forecasts started from slightly different histories should not diverge. The structural condition $q<1$ establishes exactly this behavior, and the near-zero \SI{120}{\minute} residual gain shows that almost none of an initial difference persists at the end of the horizon.

PR-RL provides these properties through one physical transition and one small residual policy network. The same heat-balance form and input definitions apply to both devices, with parameters identified for each evaluation condition. This structure turns the damping requirement into an explicit, verifiable condition ($q<1$) whenever such a contractive physical transition is extended by a restricted learned residual. For AEL systems, whose thermal dynamics motivate model-based temperature management \cite{ulleberg2003,sakas2022,qi2023}, these results provide a compact predictor that combines forecast correction with limited propagation of initial temperature-history differences.

The evaluation uses fixed public records from two devices, including records used during method development, and the confidence intervals quantify training-seed variability under these fixed records and settings. The practical judgments use the \SI{0.10}{\degreeCelsius} engineering margin, with sensitivity to other margins reported in Supplementary Note~\ref{si:margins}. The Brauns P95 comparison favors the matched RL-only model, which in turn lacks the damping of initial temperature-history differences (Fig.~\ref{fig:joint}), and the PEACE P95 comparison with the TD3-only model remains undecided because of its wide interval. Simple reference predictors such as persistence were not evaluated and would provide an additional practical accuracy baseline. Forecasts hold future inputs at their start-time values, so the predictor requires no forecasts of the exogenous inputs. Some effective thermal parameters lie at their identification bounds. The supplementary results show that operating-domain changes, transfer between devices, and longer horizons remain open challenges. In these analyses, the accuracy of the identified effective parameters outside their identification conditions appears as one important factor; because these parameters appear explicitly in the model, improved physical identification offers a direct route for extending the method. Future operating records that include cooling-system measurements would allow such richer identification and a test of PR-RL within AEL temperature-control schemes.

The results establish that physical damping and sequential error correction can be obtained together when learning is added to an AEL thermal model: the physical transition and restricted feedback provide the damping, while reward-based residual learning evaluates corrections through their effects on subsequent predictions. PR-RL therefore offers a practical design principle for temperature prediction in renewable-powered alkaline electrolyzers: the correction of forecast errors and the suppression of the influence of the initial temperature history can be designed together in one compact model.

\section{Methods}
\label{sec:methods}

\subsection{Prediction task and datasets}
\label{sec:task}

The prediction target is a representative electrolyte outlet temperature of an AEL system, predicted recursively at a time step of \SI{60}{\second}: each predicted temperature is fed back as an input for the next step, and the exogenous operating inputs are held at their values at the prediction start time. The default prediction horizon is \SI{30}{\minute}, and horizons up to \SI{120}{\minute} are used for the evaluation of initial temperature-history differences.

Two public measured operating records of different AEL devices are used. The first dataset was published by Brauns and Turek for a pressurized single-cell alkaline electrolysis system operated with 32~wt\% potassium hydroxide electrolyte \cite{brauns2023data}. Four operating records from 2021, about \SI{187}{\hour} in total at a \SI{60}{\second} logging interval, are used: two records for training, one for validation, and one for final evaluation, with the average of the anode-side and cathode-side outlet electrolyte temperatures as the target. The second dataset was published by the Pressurized Efficient Alkaline EleCtrolysEr (PEACE) project for a pressurized three-cell short stack with 30~wt\% potassium hydroxide electrolyte \cite{peace2026}. Its three experiments, about \SI{8.7}{\hour} in total and recorded at a native interval of about one second, are resampled to the common \SI{60}{\second} grid by taking the value of each variable at the end of each \SI{60}{\second} interval; one experiment each serves for training, validation, and final evaluation, and the target temperature is the average of the hydrogen-side and oxygen-side outlet electrolyte temperatures. These splits define the main condition; for the statistical analysis in the `Statistical analysis' section of Methods, the role of the evaluation session is rotated so that each session of a device serves once as the held-out evaluation session. In every rotated condition, the sessions are ordered chronologically, the latest remaining session serves as the validation session, and the other remaining sessions form the training set, which determines all seven splits uniquely.

Table~\ref{tab:datasets} summarizes the two datasets.

\begin{table}[!htbp]
\def\baselinestretch{1}\normalsize\small
\caption{\textbf{Summary of the two measured AEL operating datasets.} Both datasets are public Zenodo records and are resampled or logged on a common \SI{60}{\second} grid for the recursive temperature prediction task. For both devices, electrolyte flow rate and pressure are kept as archived variables and are not used as model inputs. The splits listed here define the main condition; in the statistical analysis (the `Statistical analysis' section of Methods) the role of the evaluation session is rotated so that each session of a device serves once as the held-out evaluation session.}
\label{tab:datasets}
\centering
\small
\begin{tabular}{@{}p{0.24\linewidth}p{0.33\linewidth}p{0.33\linewidth}@{}}
\toprule
 & Brauns & PEACE \\
\midrule
Device & Pressurized single-cell alkaline electrolysis system & Pressurized three-cell short stack \\
Electrolyte & 32~wt\% KOH & 30~wt\% KOH \\
Data source & Brauns and Turek, Zenodo \cite{brauns2023data} & PEACE project, Zenodo \cite{peace2026} \\
Records used & Four operating records from 2021 (sessions 1--4: June 19, June 22, June 29, July 5) & Three experiments (1--3) \\
Total duration & About \SI{187}{\hour} & About \SI{8.7}{\hour} \\
Native logging interval & \SI{60}{\second} & About \SI{1}{\second}, with occasional gaps (resampled to \SI{60}{\second})\textsuperscript{a} \\
Split (training\,/\allowbreak\,validation\,/\allowbreak\,evaluation) & 2\,/\,1\,/\,1 records & 1\,/\,1\,/\,1 experiments \\
Target temperature & Average of anode-side and cathode-side outlet electrolyte temperatures & Average of hydrogen-side and oxygen-side outlet electrolyte temperatures \\
Target temperature range (common grid) & 58.11--\SI{59.61}{\degreeCelsius} & 14.94--\SI{32.71}{\degreeCelsius} \\
Cell voltage $V_{\mathrm{cell}}$ & Stack voltage\,/\,$N$ with $N=1$ & Stack voltage\,/\,$N$ with $N=3$ \\
\bottomrule
\end{tabular}

\smallskip
\raggedright\footnotesize
\textsuperscript{a}\,In the resampling, each variable is taken as its value at the end of each \SI{60}{\second} interval.
\end{table}

For both devices the model input vector $x_k$ contains thirteen variables: the temperature $T$ at the present time step, the inlet electrolyte temperature $T_{\mathrm{in}}$, the electric current $I$, and the cell voltage $V_{\mathrm{cell}}$, together with the changes of $T$, $I$, and $V_{\mathrm{cell}}$ over the past 1, 5, and \SI{10}{\minute}. The cell voltage is obtained as the stack voltage divided by the number of series-connected cells $N$ ($N=1$ for the single-cell system and $N=3$ for the short stack), whereas the current of a series stack is not divided. Electrolyte flow rate and pressure are kept as archived variables and are not used as model inputs. Input normalization and the residual action range are determined from the training data only; their exact definitions are given in Supplementary Note~\ref{app:config}. Throughout, prediction windows are formed only where the measurements are complete on the common \SI{60}{\second} grid: every start point requires a complete \SI{10}{\minute} temperature history, a missing value or a logging interruption ends the usable target interval, and the accuracy evaluation uses only start points whose full target horizon is observed.

\subsection{Common physics-based thermal model}
\label{sec:physics}

PR-RL and the physics-based comparison models share one lumped thermal balance of the electrolyte outlet temperature, written per cell as
\begin{equation}
C\,\frac{\mathrm{d}T}{\mathrm{d}t} = I\left(V_{\mathrm{cell}}-1.48\right) - K\left(T-T_{\mathrm{ref}}\right) - G\left(T-T_{\mathrm{in}}\right),
\qquad C,K,G>0,
\label{eq:thermal}
\end{equation}
where \SI{1.48}{\volt} is the thermoneutral voltage, so that the first term represents the electrochemical heat generation, the second term the heat exchange with an effective boundary temperature $T_{\mathrm{ref}}$, and the third term the heat transport by the circulating electrolyte with an effective constant coefficient $G$. The per-cell heat capacity $C$ and heat transfer coefficients $K$ and $G$ correspond to the stack-level quantities divided by $N$. For inputs held constant over one \SI{60}{\second} step, equation~\eqref{eq:thermal} has the exact discrete transition
\begin{equation}
\Phi(T,u) = A\,T + (1-A)\,T_{\mathrm{eq}},
\qquad A = \exp\!\left[-\frac{60\,(K+G)}{C}\right],
\label{eq:transition}
\end{equation}
where $u$ collects the exogenous inputs and $T_{\mathrm{eq}}$ is the equilibrium temperature implied by equation~\eqref{eq:thermal} for those inputs. The parameters $C$, $K$, and $G$ are constrained to be positive, and $T_{\mathrm{ref}}$ is identified within a bounded range. All four parameters are identified from the training records of each evaluation condition by nonlinear least squares on the predicted temperatures of the transition in equation~\eqref{eq:transition}; the numerical settings of this identification are listed in Supplementary Note~\ref{app:physparams}. For the Brauns dataset, the recursive predictions with the exogenous inputs held at their start values are fitted to the measured temperatures at prediction lengths of 1, 5, 15, and \SI{30}{\minute}, matching the structure of the prediction task; for the PEACE dataset, the fit uses the measured one-step (\SI{60}{\second}) transitions of the training record. The identified parameters are then shared by the physics-based model and all residual models of that condition, so that all models that contain the physical transition start from identical identified parameters. The identified values are effective parameters of the lumped model; $T_{\mathrm{ref}}$ is an effective boundary temperature rather than an independently measured ambient temperature.

\subsection{PR-RL: reinforcement-learning residual policy}
\label{sec:prrl}

PR-RL predicts the next temperature as the physical transition plus a bounded residual produced by a single small reinforcement-learning residual policy,
\begin{equation}
\hat{T}_{k+1} = \Phi(\hat{T}_k,u_k) + a_{\max}\tanh\!\big(g_\theta(x_k)\big),
\label{eq:prrl}
\end{equation}
where $g_\theta$ is a fully connected network with two hidden layers of 64 nodes each (13--64--64--1, 5{,}121 parameters) and $a_{\max}$ is the residual action range determined from the training residuals. The residual action is a correction of the predicted state, not a command to the physical plant. No auxiliary predictor, learned critic, ensemble, model predictive controller, or post-hoc output clipping is added, so the inference structure consists of the physical transition and one residual policy network.

Fig.~\ref{fig:framework} summarizes the framework. Panel~(a) shows the inference structure of equation~\eqref{eq:prrl}: at each \SI{60}{\second} step, the thirteen-variable input vector of the `Prediction task and datasets' section of Methods is passed to the residual policy, the bounded residual is added to the physical transition, and the predicted temperature is fed back into the temperature history for the next step, with the exogenous inputs held at their values at the prediction start time. Panel~(b) shows the training signal flow described in the remainder of this section, in which rewards formed from the measured temperatures evaluate each correction through its effect on the model's own recursive predictions. Panel~(c) illustrates the damping mechanism of the `Damping of initial temperature-history differences' section of Methods, by which a difference between the initial temperature histories of two predictions under the same inputs decays along the prediction horizon.

The residual policy is trained as a deterministic policy on the model's own predicted states. A residual action changes the predicted temperature that enters the subsequent policy inputs and therefore affects the later predictions. Because the action modifies the prediction rather than any input of the physical plant, alternative residual actions can be evaluated against the same measured temperatures during training. Starting from a decision time $k$ within a training episode, the model rolls out its recursive predictions, and the action value of a normalized residual action $a$ taken at that time is estimated from the squared errors between the subsequent predicted and measured temperatures,
\begin{equation}
\hat{Q}(x_k, a) = -\frac{1}{n}\sum_{j=k}^{n-1}
\left( \frac{\hat{T}_{j+1} - T^{\mathrm{meas}}_{j+1}}{s_T} \right)^{\!2},
\label{eq:actionvalue}
\end{equation}
where $\hat{T}_{j+1}$ are the predicted temperatures of the rollout that applies the action $a$ at time $k$ and follows the current policy afterwards, $s_T$ is the temperature normalization scale, the standard deviation of the measured temperature over the training records, and $n$ is the episode length, limited to \SI{30}{\minute}; when fewer measurements remain in the record, only the actually observed length contributes, and during the early training epochs $n$ is additionally limited by the stepwise reward-horizon schedule listed in Supplementary Table~\ref{tab:config}. The gradient of the action value with respect to the action is estimated numerically from two additional rollouts in which the normalized action is shifted in opposite directions,
\begin{equation}
\frac{\partial \hat{Q}}{\partial a} \approx
\frac{\hat{Q}(x_k, a^{+}) - \hat{Q}(x_k, a^{-})}{a^{+} - a^{-}},
\qquad a^{\pm} = \operatorname{clip}(a \pm \varepsilon,\, -1,\, 1),
\label{eq:fd}
\end{equation}
with $\varepsilon = 0.002$; at the boundaries of the action range, the denominator is the actual difference between the two clipped actions. Only the policy network is updated. The training objective $J$ is the expected return of a training episode, that is, the negative mean of the normalized squared prediction errors over the whole episode, which equals the action value of equation~\eqref{eq:actionvalue} at the episode start under the current policy. Following the deterministic policy gradient theorem \cite{silver2014}, the gradient of $J$ is estimated from the action values of the visited decision times as
\begin{equation}
\hat{\nabla}_{\theta} J = \frac{1}{m}\sum_{k \in \mathcal{K}}
n_k\, \frac{\partial \hat{Q}}{\partial a}\bigg|_{x_k}
\nabla_{\theta}\, \tanh\!\big(g_\theta(x_k)\big),
\label{eq:update}
\end{equation}
here $\mathcal{K}$ is a set of $m$ decision times sampled uniformly within the observed episode lengths, and the weight $n_k$, the length of the episode containing decision time $k$, makes the sampled average estimate the sum over all decision times of each episode. In the implementation, the finite-difference factors $\partial\hat{Q}/\partial a$ enter as fixed weights, and the adaptive moment estimation (Adam) optimizer minimizes, with gradient-norm clipping, the surrogate loss $-\frac{1}{m}\sum_{k\in\mathcal{K}} n_k\,(\partial\hat{Q}/\partial a)\big|_{x_k}\tanh\!\big(g_\theta(x_k)\big)$, whose gradient is the negative of equation~\eqref{eq:update}, so the update increases the estimated return. Neither the physical model nor the rollout is differentiated, so the approach does not require a differentiable physics implementation. Learning rates, training lengths, and the number of sampled decision times per update are fixed configuration values recorded with the experiments; by default, three decision times are sampled per episode and update. The complete configuration is listed in Supplementary Table~\ref{tab:config}.

\subsection{Damping of initial temperature-history differences}
\label{sec:damping}

The physical transition in equation~\eqref{eq:transition} contracts temperature differences at a fixed rate $A<1$ per step. To preserve this property in the combined model, the influence of the residual policy on the temperature-history inputs is restricted. Let $L_\delta$ denote an upper bound on the Lipschitz constant of the residual output with respect to the temperature history, measured in the maximum norm over the eleven most recent temperature samples. Because only four network inputs depend on this history (the present temperature and its changes over the past 1, 5, and \SI{10}{\minute}), such a bound is computed directly from the network weights and the input normalization,
\begin{equation}
L_\delta = a_{\max}\, \lVert W_3 \rVert_2\, \lVert W_2 \rVert_2
\max_{v \in \{-1,1\}^4} \lVert W_1 E\, v \rVert_2,
\label{eq:ldelta}
\end{equation}
where $W_1$, $W_2$, and $W_3$ are the weight matrices of the three network layers, the matrix $E \in \mathbb{R}^{13\times 4}$ maps a perturbation of the four delayed temperature samples that enter the inputs through the input normalization to the four temperature-related network inputs and has nonzero entries only in the four corresponding input rows, and the maximization over the sign patterns $v$ covers the worst-case combination of unit perturbations of those four samples; the rectified linear unit (ReLU) and tanh activation functions have slope at most one, so the product of the layer norms bounds the change of the residual output. If
\begin{equation}
q = A + L_\delta < 1,
\label{eq:q}
\end{equation}
then for two predictions driven by the same exogenous inputs but different initial temperature histories, the maximum $M_t$ of the differences between the two predicted temperatures over the most recent eleven samples (the \SI{10}{\minute} temperature history at the \SI{60}{\second} step) satisfies $M_{t+11}\le q\,M_t$, so the initial difference decays along the prediction horizon. This follows by a short induction: the physical transition is linear in the predicted temperature with factor $A$, so the difference between the two predictions after one step satisfies $|\Delta T_{k+1}| \le A\,|\Delta T_k| + L_\delta\, M_k \le (A + L_\delta)\, M_k = q\, M_k$. Every newly generated difference is therefore at most $q\,M_t \le M_t$, so the window maximum never increases, and after eleven steps the history window contains only new values, which gives $M_{t+11} \le q\, M_t$. During training, the bound $L_\delta$ is recomputed immediately after every optimizer update, and whenever it exceeds $\rho\,(1-A)$, the four first-layer weight columns associated with the temperature inputs are scaled by the common factor $\rho\,(1-A)/L_\delta$; this restores $L_\delta \le \rho\,(1-A)$ after every update, keeps equation~\eqref{eq:q} satisfied throughout training, and leaves all other weights unchanged. The projection fraction is $\rho=0.8$ in the default configuration used for six of the seven evaluation conditions of the `Statistical analysis' section of Methods; for the remaining Brauns condition, the configuration selected on validation data uses $\rho=0.5$. The identical projection is applied to the supervised residual MLP, so the damping condition itself is a property of the model structure and is not attributed to the learning scheme. This condition concerns the difference between two predicted trajectories under the same inputs and is evaluated separately from the prediction error with respect to the measured temperature.

\subsection{Comparison models}
\label{sec:comparison}

Five comparison models are evaluated under matched conditions: the same data splits, policy inputs, prediction start points, and test conditions as PR-RL.
\begin{itemize}
\item \textbf{Physics-based model.} The recursive transition of equation~\eqref{eq:transition} alone, with the identified parameters.
\item \textbf{Physics-based model with supervised residual MLP.} A network with the identical thirteen inputs, architecture, and damping projection as the residual policy, trained by direct supervised regression of the measured one-step residual $r_k^{*} = T_{k+1}^{\mathrm{meas}} - \Phi(T_k^{\mathrm{meas}},u_k)$ with a mean squared error loss; the regression target is the normalized residual $r_k^{*}/a_{\max}$, and the network output passes through the identical $\tanh$ bound, so the trained model adds $a_{\max}\tanh(\cdot)$ to the physical transition exactly as in equation~\eqref{eq:prrl}. No reward, policy gradient, or backpropagation through recursive trajectories is used for this model, so the comparison with PR-RL isolates the effect of the learning scheme: direct one-step regression versus sequentially rewarded learning on self-predicted states.
\item \textbf{Physics-based model with Random Forest residual.} The same supervised one-step residual target learned by a Random Forest \cite{breiman2001} with 300 trees and a minimum leaf size of 3, serving as an additional non-neural reference; its normalized output is clipped to $[-1,1]$ and scaled by the same $a_{\max}$, and the weight projection defined for neural networks is not applicable to this model.
\item \textbf{Matched RL-only model.} The same deterministic policy learning applied to the full temperature change without the physical transition, so that $\hat{T}_{k+1} = \hat{T}_k + b_{\max}\tanh\!\big(g_\theta(x_k)\big)$, where the output range $b_{\max}$ is determined from the measured one-step temperature changes of the training records by the same rule that defines $a_{\max}$ (Supplementary Note~\ref{app:config}). Because the damping projection of the `Damping of initial temperature-history differences' section of Methods is defined relative to the contraction of the physical transition, it is not applicable to this model; the comparison with PR-RL therefore isolates, under an identical learning scheme, the contribution of the physics-based structure, that is, the physical transition together with the bounded and projected residual acting on it.
\item \textbf{TD3-only model.} A second predictor without the physical transition, with the same thirteen policy inputs, the same 13--64--64--1 policy-network architecture, and the same output range $b_{\max}$ as the matched RL-only model, so that $\hat{T}_{k+1} = \hat{T}_k + b_{\max}\tanh\!\big(g_\theta(x_k)\big)$. Its policy is trained with twin delayed deep deterministic policy gradient (TD3) \cite{fujimoto2018}, a representative actor--critic algorithm whose two critics are used only during training; the reward is the same negative normalized squared prediction error divided by the episode length. The critics additionally receive the elapsed episode fraction and the normalized episode length, which represent the finite training horizon, contain no future temperature measurements, and are not inputs of the policy. Evaluation uses the deterministic policy without exploration noise, and configurations and checkpoints are selected on the validation score $S$; the complete training settings are listed in Supplementary Note~\ref{app:config}. The comparison with PR-RL therefore places the physics-residual design against a widely used reinforcement-learning algorithm.
\end{itemize}
The supervised residual MLP and the Random Forest residual model share the identified physical parameters with PR-RL, and all five comparison models share the policy inputs, prediction start points, and test conditions. The neural networks are initialized by first drawing random weights for a network that additionally contains the electrolyte flow-rate input and then removing that input column, so that the exclusion of the flow-rate input does not alter the random initialization of the retained weights; in every policy or supervised residual network the last layer is initialized to zero, so the learned output is zero at initialization and a residual model starts from the pure physics-based prediction; the TD3 critics use the default linear-layer initialization.

Table~\ref{tab:models} summarizes the six models and their comparisons.

\begin{table}[!htbp]
\def\baselinestretch{1}\normalsize\footnotesize
\caption{\textbf{Comparison-model matrix.} All models that contain the physical transition $\Phi$ of equation~\eqref{eq:transition} use identical identified parameters ($C$, $K$, $G$, $T_{\mathrm{ref}}$), and all six models share the same policy inputs, prediction start points, and test conditions. For each model the matrix lists the learned component, the learning target or signal together with the gradient path, and the damping projection, and states which factor the comparison of that model with PR-RL isolates.}
\label{tab:models}
\centering
\scriptsize
\setlength{\tabcolsep}{2pt}
\begin{tabular}{@{}>{\raggedright\arraybackslash}p{1.7cm}>{\raggedright\arraybackslash}p{1.65cm}>{\raggedright\arraybackslash}p{2.35cm}>{\raggedright\arraybackslash}p{2.95cm}>{\raggedright\arraybackslash}p{1.75cm}>{\raggedright\arraybackslash}p{2.0cm}@{}}
\toprule
Model & Physical transition $\Phi$, equation~\eqref{eq:transition} & Learned component & Learning target or signal; gradient path & Damping projection\textsuperscript{a} & Comparison with PR-RL isolates \\
\midrule
PR-RL (proposed) & Shared identified parameters & Reinforcement-learning residual policy (13--64--64--1) & Reward from subsequent prediction errors on self-predicted states (up to \SI{30}{\minute}); numerical action-value gradient, policy-network backpropagation only, physics not differentiated & Applied & --- \\
\addlinespace[2pt]
Physics-based model & Shared identified parameters & None & --- & Not applicable & Effect of the residual correction (contribution of the learned residual) \\
\addlinespace[2pt]
Physics + supervised residual MLP & Shared identified parameters & Supervised residual MLP (identical 13--64--64--1) & Direct regression of the measured one-step residual $r_k^{*}$\textsuperscript{b}; backpropagation of the one-step loss, physics not differentiated & Applied (identical) & Effect of the learning scheme (sequential reward versus one-step regression) \\
\addlinespace[2pt]
Physics + Random Forest residual & Shared identified parameters & Random Forest (300 trees, minimum leaf size 3) & Direct regression of the measured one-step residual; tree ensemble, no gradients & Not applicable to trees & Non-neural reference \\
\addlinespace[2pt]
Matched RL-only model & None (no physical transition) & Policy network predicting the full temperature change (13--64--64--1) & Same reward signal as PR-RL, applied to the full temperature change; gradient path same as PR-RL & Not applicable (no physical transition) & Contribution of the physics-based structure (physical transition with bounded, projected residual) \\
\addlinespace[2pt]
TD3-only model & None (no physical transition) & Policy network predicting the full temperature change (identical 13--64--64--1); two critics used only during training & Same reward signal, optimized with TD3 (two critics, delayed policy updates, target-policy smoothing) instead of the numerical action-value gradient & Not applicable (no physical transition) & Comparison with a representative actor--critic learning algorithm \\
\midrule
\multicolumn{6}{@{}l@{}}{Policy inputs, prediction start points, and test conditions: identical for all six models (the TD3 critics receive two additional training-only time inputs).} \\
\bottomrule
\end{tabular}

\smallskip
\begin{minipage}{\textwidth}
\scriptsize
\textsuperscript{a}\,Projection of the first-layer temperature-input weight columns so that $L_\delta \le \rho\,(1-A)$.\\
\textsuperscript{b}\,$r_k^{*} = T_{k+1}^{\mathrm{meas}} - \Phi(T_k^{\mathrm{meas}},u_k)$, the measured one-step residual of the physical transition.
\end{minipage}
\end{table}

\subsection{Evaluation metrics}
\label{sec:metrics}

Prediction accuracy is evaluated on recursive predictions against the measured temperature, from prediction start points spaced \SI{5}{\minute} apart in the held-out session. The primary short-term error score $S$ is the mean of the root mean square errors at the 15 and \SI{30}{\minute} horizons. The trajectory maximum error P95 is obtained by taking, for each prediction start point, the maximum absolute error within the \SI{30}{\minute} trajectory, and then the 95th percentile of these maxima over all start points. Both metrics are expressed in \si{\degreeCelsius}.

The propagation of initial temperature-history differences is evaluated by adding \SI{0.5}{\degreeCelsius} to the initial temperature history of one of two otherwise identical predictions at every \SI{5}{\minute} start point and following both predictions under the same exogenous inputs. The dimensionless gain at time $t$ is the absolute difference between the two predicted trajectories divided by the initial \SI{0.5}{\degreeCelsius} perturbation. The start points are the same as in the accuracy evaluation, that is, the complete \SI{30}{\minute} windows at \SI{5}{\minute} spacing; because this evaluation compares two predictions with each other, it requires no measured temperatures after the prediction start, and both predictions are rolled out to \SI{120}{\minute} from every such start point. The gain is evaluated at the \SI{60}{\second} prediction steps from 1 to \SI{120}{\minute}; at the start time itself the gain equals one by construction. Two summary quantities are reported: the maximum propagation gain (the 95th percentile, over start points, of the maximum gain within each \SI{120}{\minute} trajectory) and the \SI{120}{\minute} residual gain (the 95th percentile, over start points, of the gain remaining at \SI{120}{\minute}). Gains below one indicate that the initial difference is attenuated rather than amplified.

\begin{figure}[!tp]
\def\baselinestretch{1}\normalsize
\centering
\includegraphics[width=0.96\textwidth]{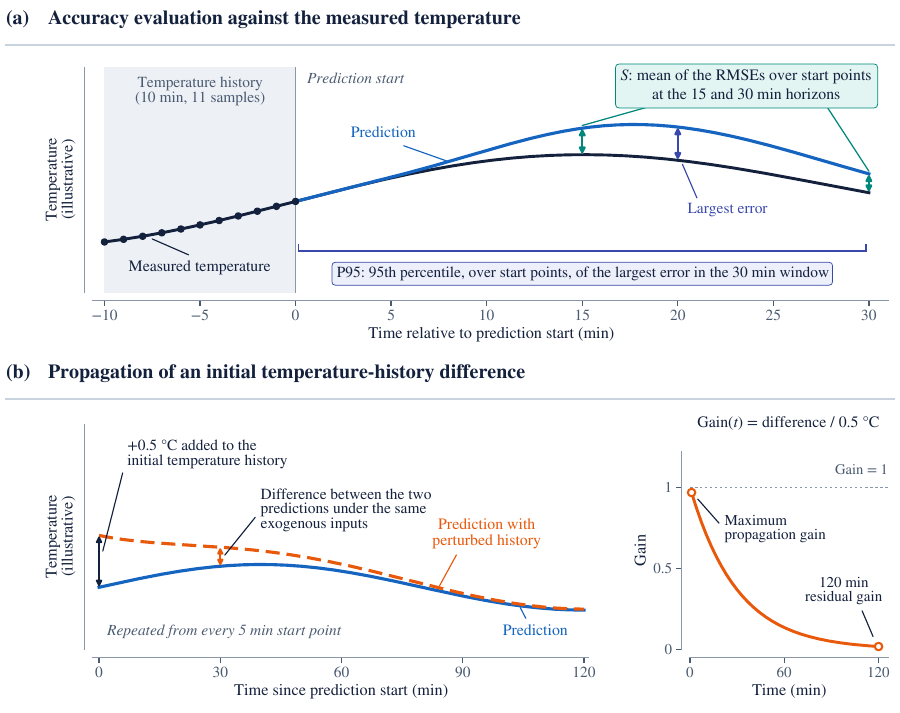}
\caption{\textbf{Evaluation protocol for recursive predictions.} Evaluation protocol for the recursive temperature predictions (the `Evaluation metrics' section of Methods), shown with illustrative curves that indicate the procedure rather than measured results. (a)~Accuracy evaluation against the measured temperature: the shaded band marks the \SI{10}{\minute} temperature history before the prediction start, the green arrows mark the errors at the 15 and \SI{30}{\minute} horizons that define $S$, and the indigo arrow and bracket mark the largest absolute error within the \SI{30}{\minute} trajectory that enters P95. (b)~Evaluation of the propagation of an initial temperature-history difference: one of two otherwise identical predictions receives \SI{0.5}{\degreeCelsius} added to its initial temperature history, and the dimensionless gain (right) is the difference between the two predictions divided by this perturbation.}
\label{fig:protocol}
\end{figure}

Fig.~\ref{fig:protocol} illustrates the two evaluations on a common time axis. Panel~(a) shows the accuracy evaluation against the measured temperature: the prediction starts from the \SI{10}{\minute} measured temperature history, and the marked errors at the 15 and \SI{30}{\minute} horizons and the largest error within the \SI{30}{\minute} trajectory are the quantities that enter $S$ and P95. Panel~(b) shows the initial-difference evaluation: the gain is read from the difference between the two predicted trajectories, so this evaluation requires no measured temperatures after the prediction start and is separate from the prediction error of panel~(a).

\subsection{Statistical analysis}
\label{sec:statistics}

Each model's configuration, candidate selection, and the acceptance margins were fixed from training and validation data before its evaluation-seed runs and were not changed after any evaluation results were seen. Three levels of choice are distinguished, and all of them use only training and validation data: the reward in equation~\eqref{eq:actionvalue} uses horizons of up to \SI{30}{\minute}; during training, the network checkpoint with the best short-term error score $S$ on the validation data is retained; and among the candidate configurations, those whose validation $S$ lay within five percent of the best candidate were short-listed, and the configuration with the lowest mean validation RMSE at prediction lengths of 60, 90, and \SI{120}{\minute} was selected for each evaluation condition, so that the configuration choice also accounts for horizons beyond the reward length without touching any evaluation data. The TD3-only model was trained with its own training budget and candidate configurations, and its configurations and checkpoints were selected on the validation score $S$ alone, likewise using only training and validation data (Supplementary Note~\ref{app:config}). Seeds used during configuration selection were excluded, and twenty new training seeds were run to completion for every condition, irrespective of intermediate results, covering seven leave-one-session-out conditions over the two devices; the two predictors without the physical transition were trained with the matched policy-gradient procedure and with TD3, respectively, under the same seed and evaluation protocol. Within each seed, the session-level metrics of each device are averaged with equal weights, and the paired difference between PR-RL and each comparison model is computed per seed and device; for each metric, smaller values are better, and the difference is defined as the PR-RL value minus the comparison-model value. The confidence interval (CI) of the mean paired difference is calculated with the $t$ distribution, with degrees of freedom equal to the number of seeds minus one. The main comparisons are grouped into families that cover all comparisons of the five comparison models for each metric type, and Bonferroni-adjusted simultaneous confidence intervals are used within each family: the ten comparisons of $S$ and the ten comparisons of P95---two datasets and five comparison models each---form two families with an error probability of 2.5\% each, so that the error-metric judgments of this cohort are jointly controlled at the 5\% level, and the eight comparisons of the propagation gains---two datasets, the two predictors without the physical transition, and the two gain metrics---form a separate family with an error probability of 5\%. Percentile bootstrap intervals from 20{,}000 seed resamples are reported alongside without altering any judgment. Because the precision of the first cohort was not sufficient to classify the PEACE comparison against the supervised residual MLP at the chosen margin, a separate confirmation cohort of 400 additional seeds was run for this single comparison; its size and settings were fixed before its results were seen, the final judgment of this comparison uses only the new 400 seeds, and its two error metrics ($S$ and P95) form a separate Bonferroni family with an error probability of 2.5\%. The sample unit throughout is the training seed under the fixed operating records and settings.

For the error metrics, practical superiority, equivalence, and inferiority are judged with a margin of $\delta$: a comparison is superior when the upper confidence limit of the difference is below $-\delta$, equivalent when the whole interval lies within $(-\delta,\delta)$, and inferior when the lower limit exceeds $\delta$; when none of these conditions holds, the comparison is left undecided. Non-inferiority is recorded separately when the upper limit is below $\delta$ \cite{lakens2017}. The working margin is $\delta=\SI{0.10}{\degreeCelsius}$, an engineering choice that assigns ten percent of the $\pm\SI{1}{\degreeCelsius}$ steady-state temperature variation reported for a controlled AEL system \cite{martinez2026} as the tolerable performance difference between models; judgments at \SI{0.05}{\degreeCelsius} and \SI{0.20}{\degreeCelsius} are preserved as sensitivity checks. The dimensionless propagation gains are judged against zero difference using the simultaneous confidence limits: superiority when the upper limit is below zero and inferiority when the lower limit is above zero.

Supplementary Table~\ref{tab:decisions} summarizes these decision rules.

\subsection{Use of AI-assisted technologies}
During the preparation of this work the authors used Claude (Anthropic) and Codex (OpenAI) in order to improve the language and readability of the manuscript and to assist with programming. After using these tools, the authors reviewed and edited the content as needed and take full responsibility for the content of the published article.

\section{Data availability}
The experimental datasets analysed in this study are publicly available on Zenodo \cite{brauns2023data,peace2026}.

\section{Author contributions}
Y.-W.K.: Conceptualization, Methodology, Resources, Software, Formal analysis, Investigation, Writing -- original draft. Y.-C.B.: Conceptualization, Supervision, Project administration, Funding acquisition, Writing -- review \& editing.

\section{Additional information}
\textbf{Competing interests:} The authors declare no competing interests.

\startsupplementary

\section*{Supplementary Information}

Supplementary Note~\ref{app:config} lists the complete training and configuration details, and Supplementary Note~\ref{app:physparams} gives the identified physical parameters and damping bounds of all evaluation conditions. Supplementary Note~\ref{si:method} summarizes the methodological rationale of PR-RL and its distinctions from related approaches, and Supplementary Note~\ref{si:stats} provides the detailed statistical evidence that supports the main results. Supplementary Notes~\ref{si:extrap}--\ref{si:long} show how prediction accuracy changes across operating domains, devices, and prediction horizons beyond those used for training and selection; these stored analyses were performed with the five models available before the TD3-only model was added, and the TD3-only model appears in the main comparisons and in Supplementary Note~\ref{si:stats}. All entries come from the fixed models and configurations, and the tables and figures report both favorable and unfavorable conditions.

\section{Training and configuration details}
\label{app:config}

Supplementary Table~\ref{tab:config} lists the training settings and statistical cohorts.

\begin{table}[!htbp]
\def\baselinestretch{1}\normalsize\footnotesize
\caption{\textbf{Fixed configuration and training details.} Fixed configuration and training details of the reinforcement-learning residual policy, the comparison models, and the statistical cohorts. The configuration values referenced in the `PR-RL: reinforcement-learning residual policy' and `Statistical analysis' sections of Methods were fixed before the corresponding statistical cohort was run. The listed learning rates and training lengths apply to all main experiments; all neural models are trained with the Adam optimizer.}
\label{tab:config}
\centering
\scriptsize
\renewcommand{\arraystretch}{0.94}
\begin{tabular}{@{}>{\raggedright\arraybackslash}p{0.42\textwidth}>{\raggedright\arraybackslash}p{0.52\textwidth}@{}}
\toprule
Item & Value \\
\midrule
\multicolumn{2}{@{}l}{\itshape Network architecture} \\
Residual policy network & fully connected, 13--64--64--1 (5{,}121 parameters) \\
Residual output bound & $a_{\max}\tanh(\cdot)$, with $a_{\max}=3\,Q_{0.99}(|r_k^{*}|)$ clipped to $[\SI{0.05}{\degreeCelsius},\,\SI{2.00}{\degreeCelsius}]$, where $Q_{0.99}(|r_k^{*}|)$ is the 99th percentile of the absolute measured one-step residuals over all consecutive \SI{60}{\second} training pairs \\
Output range of the predictors without the physical transition & $b_{\max}$ by the same rule, applied to the measured one-step temperature changes $T^{\mathrm{meas}}_{k+1}-T^{\mathrm{meas}}_{k}$ of the training records; shared by the matched RL-only and TD3-only models \\
Input normalization & present values standardized by the training mean and standard deviation of each variable; the nine change inputs divided by the standard deviation of the underlying variable, without centering; standard deviations floored at \SI{0.1}{\degreeCelsius} ($T$, $T_{\mathrm{in}}$), \SI{1}{\ampere} ($I$), and $0.05/N$\,\si{\volt} ($V_{\mathrm{cell}}$) \\
\addlinespace
\multicolumn{2}{@{}l}{\itshape Training of the residual policy} \\
Reward horizon & up to \SI{30}{\minute}; only the actually observed length contributes \\
Reward horizon schedule & \SI{1}{\minute} for epochs 1--5, \SI{5}{\minute} for epochs 6--10, \SI{15}{\minute} for epochs 11--15, and the full \SI{30}{\minute} horizon from epoch 16 \\
Training episode start points & every \SI{5}{\minute} (Brauns) and every \SI{1}{\minute} (PEACE) within the training records \\
Minibatch size & up to 128 episodes (policy) or one-step pairs (supervised MLP) per update \\
Decision times sampled per episode and update & 3 \\
Finite-difference action shift & $\pm 0.002$ in the normalized action \\
Learning rate and gradient clipping & $3\times10^{-4}$ (Adam with default moment parameters 0.9 and 0.999, no weight decay); gradient norm clipped at 10 \\
Training length & up to 400 epochs; early stopping after 50 epochs without validation improvement (not before epoch 100) \\
Damping projection & $\rho = 0.8$ (default), applied to the four temperature-related first-layer weight columns; $\rho = 0.5$ in one Brauns evaluation condition (the `Damping of initial temperature-history differences' section of Methods) \\
Checkpoint selection & network state with the best short-term error score $S$ on the validation data (improvement threshold $10^{-6}$; $10^{-8}$ for the supervised MLP) \\
Configuration selection & among candidates with validation $S$ within 5\% of the best, lowest mean validation RMSE at 60, 90, and \SI{120}{\minute} \\
\addlinespace
\multicolumn{2}{@{}l}{\itshape Training of the comparison models} \\
Supervised residual MLP & learning rate $1\times10^{-3}$, up to 500 epochs; early stopping after 40 epochs without validation improvement; identical network architecture and damping projection \\
Random Forest residual & 300 trees, minimum leaf size 3 \\
Matched RL-only model & same policy-learning settings as the residual policy; the damping projection is not applicable (no physical transition) \\
TD3-only model: networks & policy identical to the matched RL-only model (13--64--64--1, tanh output, last layer initialized to zero); two critics, each fully connected with two hidden layers of width $w$ ($w = 64$ or 128, ReLU), whose inputs are the thirteen policy inputs, the elapsed episode fraction $k/n$, the normalized episode length $n/30$, and the action (default linear-layer initialization) \\
TD3-only model: learning & discount factor 1, no bootstrapping at episode end; target-policy smoothing noise 0.2 clipped at $\pm0.5$ (normalized action); Polyak coefficient 0.005; one policy and target update per two critic updates; replay buffer 200{,}000 transitions, minibatch 256; 2{,}048 initial random-action transitions; 32 parallel environments with two critic updates per 32 collected transitions; policy and critic learning rate $3\times10^{-4}$ or $1\times10^{-4}$; exploration noise 0.1 or 0.2 (normalized action) \\
TD3-only model: training length and selection & up to 8{,}000 critic updates with validation every 500 updates; stopping after eight consecutive validations without improvement beyond 4{,}000 updates; the initial policy is also a selection candidate; development seeds 11, 23, and 37 over four candidate configurations; configurations and checkpoints selected on the validation score $S$ \\
\addlinespace
\multicolumn{2}{@{}l}{\itshape Statistical cohorts} \\
Main cohort & 20 new training seeds per condition, all run to completion; seven leave-one-session-out conditions over the two devices; configuration-selection seeds excluded \\
Confirmation cohort & 400 additional seeds, used only for the PEACE comparison against the supervised residual MLP \\
Evaluation start points & every \SI{5}{\minute} in the held-out session (accuracy and propagation-gain evaluations) \\
Practical margin & $\delta = \SI{0.10}{\degreeCelsius}$ working margin; \SI{0.05}{\degreeCelsius} and \SI{0.20}{\degreeCelsius} as sensitivity checks \\
\bottomrule
\end{tabular}
\end{table}

\section{Identified physical parameters and damping bounds}
\label{app:physparams}

Supplementary Table~\ref{tab:physparams} lists, for each of the seven leave-one-session-out conditions of the main cohort, the identified per-cell parameters of the lumped thermal balance of equation~\eqref{eq:thermal}, the residual action range $a_{\max}$, and the output range $b_{\max}$ shared by the matched RL-only and TD3-only models.

\begin{table}[!htbp]
\def\baselinestretch{1}\normalsize\footnotesize
\caption{\textbf{Identified parameters and damping bounds.} Identified per-cell parameters, residual action ranges, and damping bounds of the seven leave-one-session-out conditions of the main cohort. $C$ is the per-cell heat capacity (\si{\joule\per\kelvin}), $K$ and $G$ are the per-cell heat transfer coefficients (\si{\watt\per\kelvin}), $T_{\mathrm{ref}}$ is the effective boundary temperature (\si{\degreeCelsius}), $a_{\max}$ is the residual action range (\si{\degreeCelsius}), and $b_{\max}$ is the shared output range of the matched RL-only and TD3-only models (\si{\degreeCelsius}), all determined from the training records of each condition. $A$ is the per-step contraction factor of equation~\eqref{eq:transition}, $\rho$ the projection fraction, and $A+\rho\,(1-A)$ the projection limit of the damping bound. The $q$ columns give the range of the stored damping bound over the twenty training seeds of each condition for PR-RL and for the identically constrained supervised residual MLP; a single value indicates that the bounds coincide at the reported precision across all twenty seeds.}
\label{tab:physparams}
\centering
\footnotesize
\setlength{\tabcolsep}{4pt}
\resizebox{\textwidth}{!}{%
\begin{tabular}{@{}lccccccccccc@{}}
\toprule
Held-out session & $C$ & $K$ & $G$ & $T_{\mathrm{ref}}$ & $a_{\max}$ & $b_{\max}$ & $A$ & $\rho$ & $A+\rho(1-A)$ & $q$, PR-RL & $q$, supervised MLP \\
\midrule
Brauns, session 1 & $1.655\times10^{5}$ & 25.75 & 162.0 & 50.0 & 0.342 & 0.343 & 0.93421 & 0.8 & 0.98684 & 0.98684 & 0.93421 \\
Brauns, session 2 & $1.000\times10^{7}$ & 602.9 & 4958 & 50.0 & 0.352 & 0.353 & 0.96719 & 0.5 & 0.98359 & 0.98359 & 0.96719--0.98359 \\
Brauns, session 3 & $1.305\times10^{5}$ & 23.64 & 172.1 & 50.0 & 0.379 & 0.373 & 0.91395 & 0.8 & 0.98279 & 0.98279 & 0.91395--0.98279 \\
Brauns, session 4 & $1.305\times10^{5}$ & 23.64 & 172.1 & 50.0 & 0.379 & 0.373 & 0.91395 & 0.8 & 0.98279 & 0.98279 & 0.91395--0.98279 \\
PEACE, experiment 1 & $2.700\times10^{4}$ & 0.0352 & 88.24 & $-20.0$ & 1.088 & 2.000 & 0.82189 & 0.8 & 0.96438 & 0.96438 & 0.82189--0.96438 \\
PEACE, experiment 2 & $2.339\times10^{4}$ & 30.40 & 68.86 & 19.8 & 2.000 & 2.000 & 0.77525 & 0.8 & 0.95505 & 0.95505 & 0.77525--0.95505 \\
PEACE, experiment 3 & $2.339\times10^{4}$ & 30.40 & 68.86 & 19.8 & 2.000 & 2.000 & 0.77525 & 0.8 & 0.95505 & 0.77525--0.95505 & 0.77525--0.95505 \\
\bottomrule
\end{tabular}}
\end{table}

 It also lists the per-step contraction factor $A$ of equation~\eqref{eq:transition}, the projection fraction $\rho$, the resulting projection limit $A+\rho\,(1-A)$, and the ranges of the stored damping bound $q$ over the twenty training seeds for PR-RL and for the identically constrained supervised residual MLP.

The identification uses the following numerical settings. The positive parameters $C$, $K$, and $G$ are identified in logarithmic form, and all four parameters are bounded: $C \in [10^{2}, 10^{7}]$~\si{\joule\per\kelvin}, $K \in [10^{-5}, 10^{4}]$~\si{\watt\per\kelvin}, $G \in [10^{-4}, 10^{4}]$~\si{\watt\per\kelvin}, and $T_{\mathrm{ref}} \in [-20, 50]$~\si{\degreeCelsius}; for the three-cell stack, the bounds and initial values of $C$, $K$, and $G$ are divided by $N=3$, so that they apply to the per-cell quantities. The bounded nonlinear least-squares problem is solved with the trust-region reflective method of SciPy (\texttt{least\_squares}) with at most 160 evaluations of the residual function, started from four initial values $C \in \{10^{3}, 10^{4}, 10^{5}, 10^{6}\}$~\si{\joule\per\kelvin} with $K = 2$~\si{\watt\per\kelvin}, $G = 10$~\si{\watt\per\kelvin}, and $T_{\mathrm{ref}} = \SI{25}{\degreeCelsius}$; the solution with the lowest final cost is selected. The multistep fit of the Brauns dataset uses every fifth start point of the training records, and the one-step fit of the PEACE dataset uses all consecutive \SI{60}{\second} pairs (the `Common physics-based thermal model' section of Methods). The identified values are effective parameters of the lumped model, and several of them lie at the bounds of their admissible identification ranges, consistent with their role as effective quantities rather than directly measured physical properties.

\begin{figure}[!htbp]
\def\baselinestretch{1}\normalsize
\centering
\includegraphics[width=0.85\textwidth]{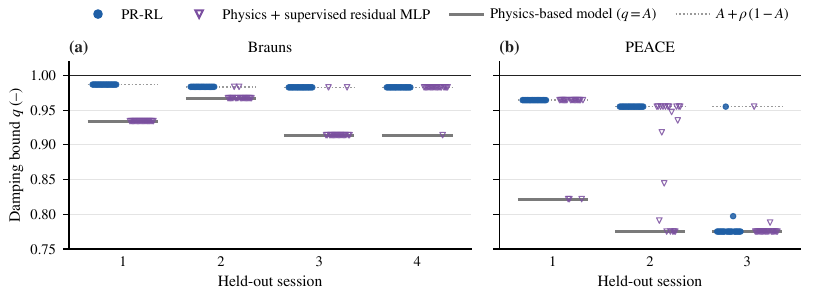}
\caption{\textbf{Stored damping bounds of the main cohort.} Stored damping bounds $q = A + L_\delta$ of PR-RL (filled circles) and the identically constrained supervised residual MLP (open triangles) for every held-out session and training seed of the main cohort, for (a)~Brauns and (b)~PEACE. Solid grey segments mark the contraction factor $A$ of the physics-based model (equation~\eqref{eq:transition}), dotted segments the projection limit $A+\rho\,(1-A)$, and the solid black line $q=1$; coincident values overlap.}
\label{fig:damping}
\end{figure}

Supplementary Fig.~\ref{fig:damping} shows all 280 stored $q$ values of the main cohort.

 The projection fraction is $\rho=0.8$ for all sessions except Brauns session~2, whose selected configuration uses $\rho=0.5$ (the `Damping of initial temperature-history differences' section of Methods). Values of $q$ equal to $A$ correspond to networks whose residual output has zero sensitivity to the temperature history, including retained initial checkpoints; values at the projection limit correspond to networks for which the projection of the `Damping of initial temperature-history differences' section of Methods was active at the selected checkpoint. All 280 values are below one, with maxima of 0.98684 for PR-RL and 0.98359 for the supervised MLP. Each $q$ value corresponds to one trained model of the main cohort under the fixed operating records.

 \clearpage

\section{Methodological rationale and distinctions from related approaches}
\label{si:method}

Supplementary Table~\ref{tab:si-method} summarizes the methodological rationale of PR-RL. For each design aspect, the table states the related approach, the corresponding PR-RL design, and the supporting evidence of this study; all judgments follow the sections, tables, and figures referenced in each row.

\begin{table}[!htbp]
\def\baselinestretch{1}\normalsize\footnotesize
\caption{\textbf{Methodological rationale of PR-RL and distinctions from related approaches.} Each row states one design aspect, the related approach, the PR-RL design with its rationale, and the contribution with its supporting evidence. The related approaches are described for the methodological comparison only; the statistical judgments refer to the comparison models evaluated in this study (the `Comparison models' section of Methods).}
\label{tab:si-method}
\centering
\scriptsize
\setlength{\tabcolsep}{3pt}
\begin{tabular}{@{}>{\raggedright\arraybackslash}p{1.9cm}>{\raggedright\arraybackslash}p{3.3cm}>{\raggedright\arraybackslash}p{3.6cm}>{\raggedright\arraybackslash}p{3.6cm}@{}}
\toprule
Methodological aspect & Related approach & PR-RL design and rationale & Contribution and supporting evidence in this study \\
\midrule
Role of the learned residual action & Residual RL adjusts the current commands of an economic MPC for PEM electrolyzer fleets \cite{nezhadkhatami2026}, following the residual RL concept of robot control \cite{johannink2019}: the learned action changes the plant input. & The residual action corrects the next predicted temperature of the AEL thermal balance; it changes the predicted temperature history that enters the subsequent policy inputs, not any input of the physical plant (the `PR-RL: reinforcement-learning residual policy' section of Methods). & Establishes the prediction-oriented role of residual learning in this study: forecast correction instead of operating-command adjustment (the `Prediction task and datasets' and `PR-RL: reinforcement-learning residual policy' sections of Methods). \\
\addlinespace[2pt]
Measurements within the prediction horizon & RL-based state estimation updates a model-based state estimate from incoming sensor measurements \cite{mowlavi2024}. & After initialization from the measured \SI{10}{\minute} temperature history, PR-RL predicts recursively without further temperature measurements within the horizon; future measured temperatures serve only as training rewards and evaluation references. & Matches the forecasting setting of this study, in which no temperature measurements arrive during the prediction horizon (the `Prediction task and datasets' and `Evaluation metrics' sections of Methods). \\
\addlinespace[2pt]
Learning signal and visited states & Supervised residual regression fits measured residuals at measured states \cite{cilibiu2026,kim2026crossover,kim2026catalyst}; models trained on one-step targets can behave differently when their own predictions are fed back \cite{venkatraman2015}. & The residual policy is trained on the model's own predicted states with a reward formed from the subsequent prediction errors (equation~\eqref{eq:actionvalue}), so each correction is evaluated together with its effects on the later predictions. & Contribution of the reinforcement-learned residual: with identical physics, inputs, architecture, and damping projection, PR-RL is practically superior to the supervised residual MLP on Brauns and practically equivalent on PEACE in $S$ and P95 (Table~\ref{tab:accuracy}). \\
\addlinespace[2pt]
Policy training procedure & Actor--critic algorithms such as TD3 fit critic networks to guide the policy update \cite{fujimoto2018}. & The scalar residual action is evaluated directly from perturbed rollouts against the same measured temperatures, and finite differences estimate the action-value gradient (equations~\eqref{eq:actionvalue}--\eqref{eq:update}); no critic is fitted, and the physical transition is not differentiated. & The matched RL-only model shares this procedure and isolates the physics-based structure; the TD3-only model extends the comparison to a representative actor--critic algorithm (the `Comparison models' section of Methods). \\
\addlinespace[2pt]
Damping of initial temperature-history differences & Contraction conditions constrain learned recurrent models \cite{revay2020}. & The identified thermal transition supplies the contraction factor $A<1$, and the projection limits the sensitivity $L_\delta$ of the residual to the temperature history ($L_\delta \le \rho\,(1-A)$), so that $q = A + L_\delta < 1$ holds throughout training (the `Damping of initial temperature-history differences' section of Methods). & Contribution of the physical transition: all 140 main-cohort PR-RL models and all 140 identically constrained supervised MLP models satisfy the condition (Supplementary Note~\ref{app:physparams}), so the damping is a structural property of the design. \\
\addlinespace[2pt]
Joint accuracy and damping & A predictor without the physical transition learns the full temperature change and is not covered by the damping condition (the `Comparison models' section of Methods). & One physical transition and one small residual policy provide forecast correction and damping together, and both properties are evaluated with the same trained weights in every condition and seed. & Contribution of the combination: non-inferior $S$ at the \SI{0.10}{\degreeCelsius} margin and a lower \SI{120}{\minute} residual gain relative to both predictors without the physical transition on both datasets (Fig.~\ref{fig:joint}, Tables~\ref{tab:gains} and~\ref{tab:accuracy}). \\
\bottomrule
\end{tabular}
\end{table}

\section{Detailed statistical material}
\label{si:stats}

Supplementary Table~\ref{tab:decisions} lists the decision rules that are applied to each paired model comparison (Methods).

\begin{table}[!htbp]
\def\baselinestretch{1}\normalsize\small
\caption{\textbf{Decision rules applied to each paired model comparison.} For every metric, the difference is defined as $\Delta = \text{metric(PR-RL)} - \text{metric(comparison model)}$, where smaller metric values are better, so negative differences favor PR-RL. $[L, U]$ denotes the two-sided Bonferroni-adjusted simultaneous confidence interval of the mean of $\Delta$, at the family error probability assigned in the `Statistical analysis' section of Methods, and $\delta$ is the practical margin that separates negligible from meaningful performance differences.}
\label{tab:decisions}
\centering
\small
\begin{tabular}{@{}ll@{}}
\toprule
Judgment & Decision rule \\
\midrule
\multicolumn{2}{@{}l}{\itshape Error metrics $S$ and P95, margin $\delta$\textsuperscript{a}} \\
Practical superiority & $U < -\delta$ \\
Practical equivalence & $L > -\delta$ \ and \ $U < \delta$ \\
Practical inferiority & $L > \delta$ \\
Undecided & none of the above \\
\addlinespace
Non-inferiority (recorded separately) & $U < \delta$ \\
Statistical superiority & $U < 0$ \\
\midrule
\multicolumn{2}{@{}l}{\itshape Dimensionless propagation gains, judged against zero difference\textsuperscript{b}} \\
Statistical superiority & $U < 0$ \\
Statistical inferiority & $L > 0$ \\
\bottomrule
\end{tabular}
\par\medskip
\begin{minipage}{\columnwidth}
\footnotesize
\textsuperscript{a}\,The working margin $\delta = \SI{0.10}{\degreeCelsius}$ is an engineering choice (the `Statistical analysis' section of Methods); judgments at $\SI{0.05}{\degreeCelsius}$ and $\SI{0.20}{\degreeCelsius}$ are preserved as sensitivity checks.
\textsuperscript{b}\,No temperature margin is applied to the dimensionless gains. The Bonferroni adjustment is applied within each family of main comparisons: the ten comparisons of $S$, the ten comparisons of P95, and the eight comparisons of the propagation gains.
\end{minipage}
\end{table}

\subsection{Session-level paired differences}
\label{si:sessions}

\begin{figure}[!htbp]
\def\baselinestretch{1}\normalsize
\centering
\includegraphics[width=0.9\textwidth]{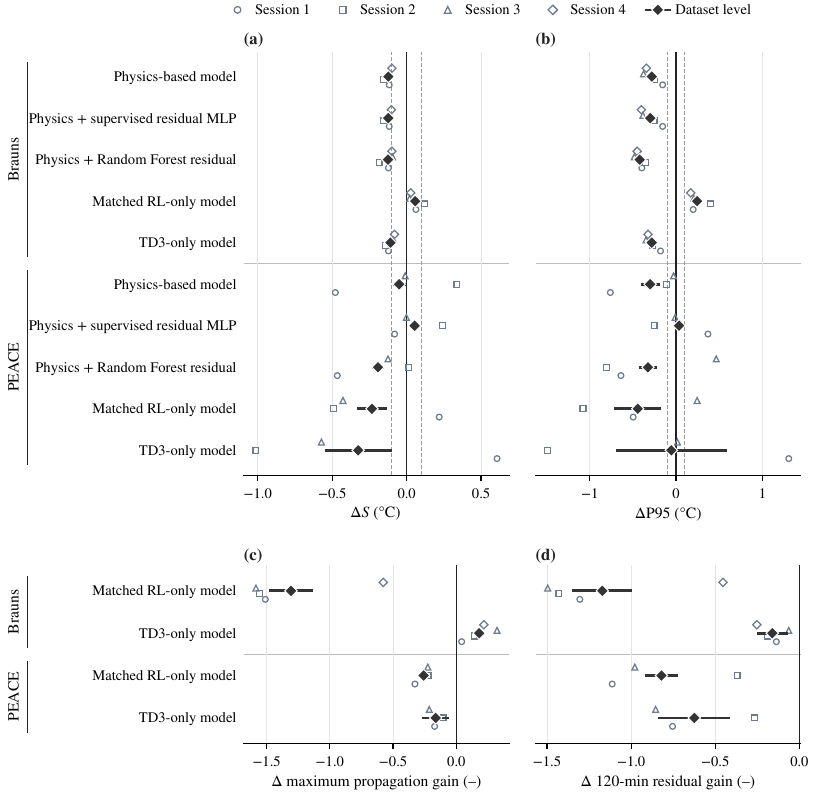}
\caption{\textbf{Session-level and dataset-level paired differences.} Session-level and dataset-level paired differences between PR-RL and each comparison model in (a)~the score $S$, (b)~P95, (c)~the maximum propagation gain, and (d)~the \SI{120}{\minute} residual gain (panels c and d: versus the matched RL-only and TD3-only models). Differences are PR-RL minus the comparison model; negative values favor PR-RL. Small open markers are session-level mean paired differences over training seeds; filled diamonds and bars are the dataset-level differences with their family-adjusted simultaneous confidence intervals, some narrower than the marker. Dashed lines in (a,~b) mark $\pm\SI{0.10}{\degreeCelsius}$.}
\label{fig:si-sessions}
\end{figure}

Supplementary Fig.~\ref{fig:si-sessions} shows the paired differences for each held-out session in every main-text comparison.

 Small session markers show the mean paired difference over training seeds for each held-out session; their spread reflects differences between the fixed operating records and is not a confidence interval. Session numbers refer to different records in the two datasets (Brauns has four sessions, PEACE three). Statistical judgments apply to the dataset-level comparisons. The dataset-level markers and intervals are those of the final comparisons in the Results section; the PEACE comparison with the supervised residual MLP uses the 400-seed confirmation cohort, and all other rows use the 20-seed main cohort. All twenty error comparisons and all eight gain comparisons are shown.

\subsection{The two PEACE cohorts of the supervised-MLP comparison}
\label{si:cohorts}

Supplementary Table~\ref{tab:si-cohorts} separates the two cohorts of the PEACE comparison against the supervised residual MLP.

\begin{table}[!htbp]
\def\baselinestretch{1}\normalsize\footnotesize
\caption{\textbf{PEACE comparison against the supervised residual MLP.} The PEACE comparison against the supervised residual MLP in the first cohort and in the confirmation cohort, in \si{\degreeCelsius}. Differences are PR-RL minus the supervised MLP with the stored simultaneous confidence intervals of the respective family; the final judgment uses only the confirmation cohort.}
\label{tab:si-cohorts}
\centering
\footnotesize
\resizebox{\textwidth}{!}{%
\begin{tabular}{@{}llcccccc@{}}
\toprule
Metric & Cohort & Seeds & PR-RL & Supervised MLP & Difference & Simultaneous CI & Practical judgment \\
\midrule
$S$ & First & 20 & 3.7616 & 3.6813 & $+0.0802$ & $[+0.0052,\,+0.1553]$ & Undecided \\
$S$ & Confirmation & 400 & 3.7572 & 3.7040 & $+0.0532$ & $[+0.0387,\,+0.0677]$ & Equivalent \\
P95 & First & 20 & 8.4861 & 8.4244 & $+0.0617$ & $[-0.1352,\,+0.2587]$ & Undecided \\
P95 & Confirmation & 400 & 8.4800 & 8.4460 & $+0.0339$ & $[-0.0027,\,+0.0706]$ & Equivalent \\
\bottomrule
\end{tabular}}
\end{table}

 In the first cohort of twenty seeds, both metrics remained undecided at the \SI{0.10}{\degreeCelsius} margin, so a separate confirmation cohort of 400 additional seeds was run for this comparison. Its size and settings were fixed before its results were seen. The final judgment uses only the new 400 seeds, and the two cohorts are not merged.

\subsection{Margin sensitivity and unadjusted intervals}
\label{si:margins}

\begin{figure}[!htbp]
\def\baselinestretch{1}\normalsize
\centering
\includegraphics[width=0.85\textwidth]{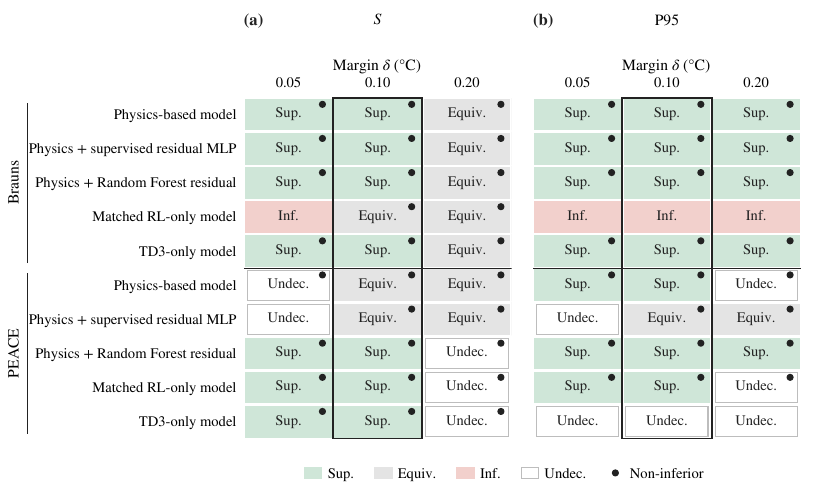}
\caption{\textbf{Sensitivity of the practical judgments to the error margin.} Practical judgments for (a)~$S$ and (b)~P95 at error margins $\delta$ of 0.05, 0.10, and \SI{0.20}{\degreeCelsius}, for all ten comparisons of each metric. Sup., Equiv., Inf., and Undec. abbreviate the superior, equivalent, inferior, and undecided judgments of the `Statistical analysis' section of Methods; dots mark non-inferiority, recorded separately from the four judgments. The framed column is the \SI{0.10}{\degreeCelsius} working margin of the main text.}
\label{fig:si-margins}
\end{figure}

Supplementary Fig.~\ref{fig:si-margins} shows the practical judgments of all twenty error comparisons at margins of 0.05, 0.10, and \SI{0.20}{\degreeCelsius}, applied to the stored family-adjusted simultaneous intervals without re-estimation; the PEACE comparison with the supervised residual MLP uses the 400-seed confirmation cohort, and all other rows use the 20-seed main cohort.

 At \SI{0.05}{\degreeCelsius}, the PEACE comparisons against the physics-based model ($S$) and against the supervised residual MLP (both metrics) become undecided, and the Brauns $S$ comparison against the matched RL-only model becomes practically inferior; the PEACE P95 comparison against the TD3-only model remains undecided at all three margins. At \SI{0.20}{\degreeCelsius}, several superior comparisons of both metrics become equivalent or undecided because the superiority condition $U<-\delta$ becomes stricter as the margin grows. The statistical direction, which is judged against zero, is unaffected by the margin. The \SI{0.10}{\degreeCelsius} working margin is the engineering tolerance for differences between model errors defined in the `Statistical analysis' section of Methods.

Supplementary Table~\ref{tab:si-intervals} lists, for the twenty-eight final comparisons of the main text, the unadjusted two-sided 95\% $t$ intervals and the percentile bootstrap intervals from 20{,}000 seed resamples that are stored alongside the simultaneous intervals.

\begin{table}[!htbp]
\def\baselinestretch{1}\normalsize\footnotesize
\caption{\textbf{Mean paired differences and intervals.} Mean paired differences of the twenty-eight final comparisons with the stored unadjusted 95\% $t$ intervals, percentile bootstrap intervals (20{,}000 seed resamples), and Bonferroni-adjusted simultaneous intervals. Error metrics are in \si{\degreeCelsius}; gains are dimensionless. The PEACE comparisons against the supervised residual MLP use the 400-seed confirmation cohort.}
\label{tab:si-intervals}
\centering
\footnotesize
\setlength{\tabcolsep}{4pt}
\resizebox{\textwidth}{!}{%
\begin{tabular}{@{}llcccc@{}}
\toprule
Dataset, comparison model & Metric & Difference & Unadjusted 95\% $t$ & Bootstrap & Simultaneous \\
\midrule
Brauns, physics-based model & $S$ & $-0.1229$ & $[-0.1253,\,-0.1205]$ & $[-0.1250,\,-0.1206]$ & $[-0.1269,\,-0.1190]$ \\
Brauns, physics + supervised MLP & $S$ & $-0.1233$ & $[-0.1260,\,-0.1206]$ & $[-0.1257,\,-0.1208]$ & $[-0.1278,\,-0.1188]$ \\
Brauns, physics + Random Forest & $S$ & $-0.1252$ & $[-0.1276,\,-0.1229]$ & $[-0.1273,\,-0.1230]$ & $[-0.1291,\,-0.1214]$ \\
Brauns, matched RL-only model & $S$ & $+0.0573$ & $[+0.0531,\,+0.0614]$ & $[+0.0536,\,+0.0610]$ & $[+0.0504,\,+0.0642]$ \\
Brauns, TD3-only model & $S$ & $-0.1089$ & $[-0.1130,\,-0.1048]$ & $[-0.1127,\,-0.1052]$ & $[-0.1157,\,-0.1021]$ \\
PEACE, physics-based model & $S$ & $-0.0506$ & $[-0.0766,\,-0.0246]$ & $[-0.0739,\,-0.0268]$ & $[-0.0938,\,-0.0074]$ \\
PEACE, physics + supervised MLP & $S$ & $+0.0532$ & $[+0.0418,\,+0.0645]$ & $[+0.0421,\,+0.0645]$ & $[+0.0387,\,+0.0677]$ \\
PEACE, physics + Random Forest & $S$ & $-0.1924$ & $[-0.2185,\,-0.1663]$ & $[-0.2157,\,-0.1683]$ & $[-0.2357,\,-0.1490]$ \\
PEACE, matched RL-only model & $S$ & $-0.2331$ & $[-0.2924,\,-0.1738]$ & $[-0.2889,\,-0.1805]$ & $[-0.3317,\,-0.1345]$ \\
PEACE, TD3-only model & $S$ & $-0.3260$ & $[-0.4609,\,-0.1910]$ & $[-0.4489,\,-0.2001]$ & $[-0.5505,\,-0.1014]$ \\
\addlinespace
Brauns, physics-based model & P95 & $-0.2811$ & $[-0.2874,\,-0.2748]$ & $[-0.2869,\,-0.2753]$ & $[-0.2916,\,-0.2706]$ \\
Brauns, physics + supervised MLP & P95 & $-0.2992$ & $[-0.3066,\,-0.2918]$ & $[-0.3061,\,-0.2924]$ & $[-0.3116,\,-0.2868]$ \\
Brauns, physics + Random Forest & P95 & $-0.4218$ & $[-0.4284,\,-0.4151]$ & $[-0.4277,\,-0.4155]$ & $[-0.4328,\,-0.4107]$ \\
Brauns, matched RL-only model & P95 & $+0.2432$ & $[+0.2317,\,+0.2548]$ & $[+0.2325,\,+0.2534]$ & $[+0.2240,\,+0.2624]$ \\
Brauns, TD3-only model & P95 & $-0.2821$ & $[-0.3036,\,-0.2605]$ & $[-0.3021,\,-0.2627]$ & $[-0.3179,\,-0.2462]$ \\
PEACE, physics-based model & P95 & $-0.3006$ & $[-0.3665,\,-0.2347]$ & $[-0.3594,\,-0.2382]$ & $[-0.4102,\,-0.1911]$ \\
PEACE, physics + supervised MLP & P95 & $+0.0339$ & $[+0.0053,\,+0.0626]$ & $[+0.0056,\,+0.0624]$ & $[-0.0027,\,+0.0706]$ \\
PEACE, physics + Random Forest & P95 & $-0.3260$ & $[-0.3895,\,-0.2625]$ & $[-0.3825,\,-0.2660]$ & $[-0.4316,\,-0.2204]$ \\
PEACE, matched RL-only model & P95 & $-0.4440$ & $[-0.6083,\,-0.2796]$ & $[-0.6072,\,-0.3077]$ & $[-0.7173,\,-0.1706]$ \\
PEACE, TD3-only model & P95 & $-0.0559$ & $[-0.4440,\,+0.3321]$ & $[-0.4112,\,+0.3053]$ & $[-0.7014,\,+0.5895]$ \\
\addlinespace
Brauns, matched RL-only model & Maximum propagation gain & $-1.3071$ & $[-1.4267,\,-1.1875]$ & $[-1.4239,\,-1.2043]$ & $[-1.4827,\,-1.1315]$ \\
Brauns, matched RL-only model & \SI{120}{\minute} residual gain & $-1.1711$ & $[-1.2931,\,-1.0492]$ & $[-1.2921,\,-1.0687]$ & $[-1.3502,\,-0.9920]$ \\
Brauns, TD3-only model & Maximum propagation gain & $+0.1807$ & $[+0.1543,\,+0.2070]$ & $[+0.1567,\,+0.2051]$ & $[+0.1420,\,+0.2193]$ \\
Brauns, TD3-only model & \SI{120}{\minute} residual gain & $-0.1607$ & $[-0.2228,\,-0.0985]$ & $[-0.2199,\,-0.1059]$ & $[-0.2520,\,-0.0694]$ \\
PEACE, matched RL-only model & Maximum propagation gain & $-0.2587$ & $[-0.2792,\,-0.2382]$ & $[-0.2788,\,-0.2412]$ & $[-0.2887,\,-0.2286]$ \\
PEACE, matched RL-only model & \SI{120}{\minute} residual gain & $-0.8200$ & $[-0.8860,\,-0.7540]$ & $[-0.8826,\,-0.7626]$ & $[-0.9169,\,-0.7231]$ \\
PEACE, TD3-only model & Maximum propagation gain & $-0.1636$ & $[-0.2365,\,-0.0907]$ & $[-0.2395,\,-0.1108]$ & $[-0.2707,\,-0.0565]$ \\
PEACE, TD3-only model & \SI{120}{\minute} residual gain & $-0.6244$ & $[-0.7704,\,-0.4785]$ & $[-0.7626,\,-0.4960]$ & $[-0.8388,\,-0.4101]$ \\
\bottomrule
\end{tabular}}
\end{table}

 The main judgments use the simultaneous intervals; the unadjusted and bootstrap intervals provide additional interval estimates for the mean paired differences.

\subsection{Distributions of the selected checkpoints}
\label{si:checkpoints}

\begin{figure}[!htbp]
\def\baselinestretch{1}\normalsize
\centering
\includegraphics[width=0.95\textwidth]{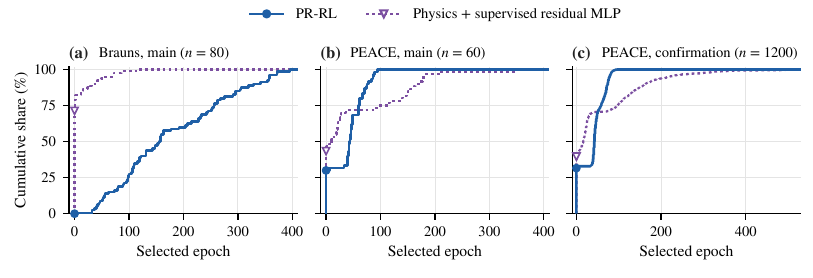}
\caption{\textbf{Training epochs of the selected checkpoints.} Empirical cumulative distributions of the training epoch of the checkpoint selected on validation data for PR-RL and for the supervised residual MLP: (a)~Brauns main cohort (4 sessions $\times$ 20 seeds, $n=80$ runs), (b)~PEACE main cohort ($3\times20$, $n=60$), and (c)~PEACE confirmation cohort ($3\times400$, $n=1200$). Markers at epoch~0 show the share of runs in which the initial zero-residual network, which reproduces the physics-based model, was retained (Supplementary Table~\ref{tab:si-epoch0}).}
\label{fig:si-checkpoints}
\end{figure}

Supplementary Fig.~\ref{fig:si-checkpoints} shows the empirical cumulative distributions of the training epoch of the checkpoint selected on validation data.

Supplementary Table~\ref{tab:si-epoch0} lists the shares of runs in which the initial (epoch-0) network was retained.

\begin{table}[!htbp]
\def\baselinestretch{1}\normalsize\small
\caption{\textbf{Shares of runs that retained the initial checkpoint.} Shares of runs in which the initial (epoch-0) checkpoint was retained by the validation-based selection. Each run is one session-exclusion condition and one training seed.}
\label{tab:si-epoch0}
\centering
\small
\begin{tabular}{@{}lccc@{}}
\toprule
Cohort & Runs & Supervised MLP & PR-RL \\
\midrule
Brauns, main & 80 & 57 (71.25\%) & 0 (0\%) \\
PEACE, main & 60 & 26 (43.33\%) & 18 (30.00\%) \\
PEACE, confirmation & 1{,}200 & 475 (39.58\%) & 379 (31.58\%) \\
\bottomrule
\end{tabular}
\end{table}

 Each run is one session-exclusion condition and one training seed, and the first PEACE cohort and the 400-seed confirmation cohort are kept separate. Retention of the initial checkpoint means that no updated candidate improved the validation score $S$; it is not interpreted as convergence or training failure. Because the two methods differ in objective function and allowed training length, the epoch distributions describe checkpoint selection rather than learning speed. The TD3-only model is trained in critic updates rather than epochs, so it is not included in these epoch distributions; its initial policy was selected in 4 of 80 Brauns runs (5.0\%) and in 32 of 60 PEACE runs (53.3\%), which, as above, is not interpreted as convergence or training failure. Performance judgments follow Table~\ref{tab:accuracy}.

\section{Operating-domain extrapolation}
\label{si:extrap}

This section reports the temperature- and load-extrapolation results. The prediction task, the score $S$, and P95 are the same as defined in the `Evaluation metrics' section of Methods. The tables average the metrics with equal weights over sessions, conditions, and training seeds. The tables provide descriptive comparisons; overlapping prediction windows and training seeds are not counted as independent devices.

In the load-extrapolation analysis of the Brauns records, contiguous segments with current $I\le\SI{50.5}{\ampere}$ form the training and validation domain and segments with $I\ge\SI{60.5}{\ampere}$ form the test domain. Both the \SI{10}{\minute} history and the \SI{30}{\minute} target of a start point must lie inside the respective domain. In the strict out-of-range evaluation of the leave-one-session-out conditions, only start points whose complete history and target lie outside the training temperature range are used.

Supplementary Table~\ref{tab:si-extrap} lists the stored results.

\begin{table}[!htbp]
\def\baselinestretch{1}\normalsize\footnotesize
\caption{\textbf{Stored load-extrapolation and strict out-of-range temperature results.} Runs count combinations of evaluation conditions and training seeds; start points are totals per seed. Entries are the score $S$ and P95 in \si{\degreeCelsius}, averaged with equal weights over sessions, conditions, and seeds.}
\label{tab:si-extrap}
\centering
\footnotesize
\setlength{\tabcolsep}{5pt}
\resizebox{\textwidth}{!}{%
\begin{tabular}{@{}lllcccc@{}}
\toprule
Analysis & Projection & Model & Runs & Start points per seed & $S$ & P95 \\
\midrule
Low-to-high load, Brauns & Not applied & PR-RL & 12 & 384 & 0.3103 & 0.4830 \\
 & & Physics-based model & 12 & 384 & 0.3115 & 0.4850 \\
 & & Physics + supervised MLP & 12 & 384 & 0.5274 & 0.7115 \\
 & & Physics + Random Forest & 12 & 384 & 0.2756 & 0.4627 \\
 & & Matched RL-only model & 12 & 384 & 0.2501 & 0.4316 \\
Low-to-high load, Brauns & Applied & PR-RL & 12 & 384 & 0.3311 & 0.5063 \\
 & & Physics-based model & 12 & 384 & 0.2628 & 0.4379 \\
 & & Physics + supervised MLP & 12 & 384 & 0.4787 & 0.7497 \\
\addlinespace
Out-of-range temperature, Brauns & Not applied & PR-RL & 3 & 2 & 0.1261 & 0.1803 \\
 & & Physics-based model & 3 & 2 & 0.3706 & 0.4176 \\
 & & Physics + supervised MLP & 3 & 2 & 0.3706 & 0.4176 \\
 & & Physics + Random Forest & 3 & 2 & 0.0589 & 0.1077 \\
 & & Matched RL-only model & 3 & 2 & 0.1243 & 0.1762 \\
Out-of-range temperature, Brauns & Applied & PR-RL & 3 & 2 & 0.1058 & 0.1640 \\
 & & Physics-based model & 3 & 2 & 0.3706 & 0.4176 \\
 & & Physics + supervised MLP & 3 & 2 & 0.3706 & 0.4176 \\
\addlinespace
Out-of-range temperature, PEACE & Not applied & PR-RL & 3 & 1 & 2.5020 & 2.7212 \\
 & & Physics-based model & 3 & 1 & 4.2788 & 4.4361 \\
 & & Physics + supervised MLP & 3 & 1 & 6.0263 & 6.1870 \\
 & & Physics + Random Forest & 3 & 1 & 4.2917 & 4.4239 \\
 & & Matched RL-only model & 3 & 1 & 2.1542 & 2.5534 \\
Out-of-range temperature, PEACE & Applied & PR-RL & 3 & 1 & 2.9212 & 3.1109 \\
 & & Physics-based model & 3 & 1 & 4.2788 & 4.4361 \\
 & & Physics + supervised MLP & 3 & 1 & 4.2950 & 4.4446 \\
\bottomrule
\end{tabular}}
\end{table}

 Configurations with and without the damping projection were selected separately and may use different identified physical parameters. Their comparison therefore includes these parameter differences as well as the projection; the projection rows contain only the evaluated physics-based models.

An additional strict temperature split separates a low training and validation domain from a high test domain at the 40th and 70th percentiles of the source training temperatures, with raw segments cut first so that both the history and the target lie inside the respective domain. All three possible Brauns splits and the one possible PEACE split were run; all three seeds use the same test temperature trajectory.

Supplementary Table~\ref{tab:si-tempsplit} lists the score $S$ of four variants that differ in their RL training rollouts and physics identification.

\begin{table}[!htbp]
\def\baselinestretch{1}\normalsize\footnotesize
\caption{\textbf{Strict temperature-split results of four variants.} Strict temperature-split results of four stored identification and training variants: score $S$ in \si{\degreeCelsius}, averaged over splits and the three seeds. The variants differ in the inputs that drive the RL training rollouts (held at start values as in deployment, measured future inputs, or both with equally weighted learning updates) and in the physics identification (as in the main analysis, or with the effective boundary temperature fixed at a nominal \SI{25}{\degreeCelsius}).}
\label{tab:si-tempsplit}
\centering
\footnotesize
\setlength{\tabcolsep}{4pt}
\resizebox{\textwidth}{!}{%
\begin{tabular}{@{}lllcccccc@{}}
\toprule
Training inputs & Physics identification & Dataset & Start points per seed & PR-RL & Physics & Supervised MLP & Random Forest & Matched RL-only \\
\midrule
Held at start values & As in the main analysis & Brauns & 90 & 1.7154 & 1.6767 & 1.7104 & 1.4309 & 0.1458 \\
 & & PEACE & 2 & 3.2269 & 3.3336 & 3.4048 & 3.4645 & 2.2997 \\
Measured future inputs & As in the main analysis & Brauns & 90 & 1.7189 & 1.6767 & 1.7104 & 1.4309 & 0.1683 \\
 & & PEACE & 2 & 3.1971 & 3.3336 & 3.4048 & 3.4645 & 2.2997 \\
Equal mixture & As in the main analysis & Brauns & 90 & 1.7490 & 1.6767 & 1.7104 & 1.4309 & 0.1550 \\
 & & PEACE & 2 & 3.1615 & 3.3336 & 3.4048 & 3.4645 & 2.2997 \\
Held at start values & Fixed \SI{25}{\degreeCelsius} boundary & Brauns & 90 & 0.8587 & 0.3932 & 0.6523 & 0.2027 & 0.1458 \\
 & & PEACE & 2 & 9.9393 & 10.0161 & 6.3470 & 10.0752 & 2.2997 \\
\bottomrule
\end{tabular}}
\end{table}

 RL training uses rollouts with exogenous inputs held at their start values as in deployment, rollouts with measured future inputs, or equally weighted learning updates from both types of rollout. The physical parameters are identified as in the main analysis (the `Common physics-based thermal model' section of Methods) or with the effective boundary temperature fixed at a nominal \SI{25}{\degreeCelsius} from the original authors' setting.

Temperature extrapolation showed no consistent advantage for PR-RL: the physical model identified in the narrower low-temperature domain was inaccurate in the higher domain, and the learned residual did not fully compensate for its errors. For PEACE, the two available test start points are also too few to judge a wide operating domain.

\section{Device transfer without adaptation}
\label{si:transfer}

In this analysis, the physical parameters, input normalization, residual ranges, and policies were determined from the source device only and were not re-estimated on the target device; the known initial \SI{10}{\minute} input history of the target supplies the prediction inputs defined in the `Prediction task and datasets' section of Methods.

Supplementary Table~\ref{tab:si-transfer} lists the bidirectional results of the main source models, averaged over the three seeds and all target sessions.

\begin{table}[!htbp]
\def\baselinestretch{1}\normalsize\small
\caption{\textbf{Bidirectional device transfer of the main source models.} Bidirectional device transfer without adaptation for the main source models: score $S$ and P95 in \si{\degreeCelsius}, averaged over all target sessions and three training seeds.}
\label{tab:si-transfer}
\centering
\small
\begin{tabular}{@{}llcc@{}}
\toprule
Transfer direction & Model & $S$ & P95 \\
\midrule
Brauns $\to$ PEACE & PR-RL & 4.7002 & 9.1554 \\
 & Physics-based model & 3.5202 & 6.7272 \\
 & Physics + supervised MLP & 4.2286 & 8.7646 \\
 & Physics + Random Forest & 3.5208 & 6.7337 \\
 & Matched RL-only model & 9.4041 & 17.1314 \\
\addlinespace
PEACE $\to$ Brauns & PR-RL & 10.7305 & 11.2613 \\
 & Physics-based model & 10.6563 & 11.1867 \\
 & Physics + supervised MLP & 11.8396 & 12.3771 \\
 & Physics + Random Forest & 10.4841 & 10.9660 \\
 & Matched RL-only model & 1.2045 & 2.1971 \\
\bottomrule
\end{tabular}
\end{table}

Supplementary Table~\ref{tab:si-transfer-cand} lists the stored score $S$ of representative additional configurations under the same no-adaptation protocol.

\begin{table}[!htbp]
\def\baselinestretch{1}\normalsize\footnotesize
\caption{\textbf{Device transfer of representative stored configurations.} Device transfer without adaptation for representative stored configurations: score $S$ in \si{\degreeCelsius}, averaged over all target sessions and three seeds. The configurations differ in the inputs driving the training rollouts, the physics identification, and the inputs used for validation-based selection; the hybrid selection combines 75\% of the validation score $S$ evaluated with measured future inputs and 25\% of the score evaluated with inputs held at start values.}
\label{tab:si-transfer-cand}
\centering
\footnotesize
\setlength{\tabcolsep}{4pt}
\resizebox{\textwidth}{!}{%
\begin{tabular}{@{}lllcccccc@{}}
\toprule
 & & & \multicolumn{3}{c}{Brauns $\to$ PEACE} & \multicolumn{3}{c}{PEACE $\to$ Brauns} \\
\cmidrule(lr){4-6}\cmidrule(l){7-9}
Training inputs & Physics identification & Validation inputs & PR-RL & Physics & Matched RL-only & PR-RL & Physics & Matched RL-only \\
\midrule
Held at start values & As in the main analysis & Held & 4.7002 & 3.5202 & 9.4041 & 10.7305 & 10.6563 & 1.2045 \\
Measured future inputs & As in the main analysis & Held & 4.2840 & 3.5202 & 5.7662 & 10.9707 & 10.6563 & 24.2222 \\
Measured future inputs & Multi-step fit on measured inputs & Measured & 6.8746 & 6.3878 & 7.2830 & 2.4602 & 6.3915 & 22.3208 \\
Measured future inputs & Multi-step fit on measured inputs & Hybrid & 6.8746 & 6.3878 & 8.8790 & 2.7350 & 6.3915 & 22.6533 \\
Held at start values & Fixed \SI{25}{\degreeCelsius} boundary, measured inputs & Held & 7.4733 & 7.1521 & 9.4041 & 1.3946 & 1.1151 & 1.2045 \\
\bottomrule
\end{tabular}}
\end{table}

 The training rollouts use either inputs held at start values as in deployment or measured future inputs. The physical parameters are identified as in the main analysis, by a multi-step fit driven by measured inputs, or with the effective boundary temperature fixed at a nominal \SI{25}{\degreeCelsius} from the original authors' setting, which is a prior value rather than a measured ambient temperature. Validation-based selection uses held inputs, measured inputs, or a hybrid score that weights the measured-input validation score $S$ by 75\% and the held-input score by 25\%. Several configurations reduce the PEACE$\to$Brauns PR-RL error, but their gains depend on the transfer direction and the identified physical model; for example, identification with the fixed \SI{25}{\degreeCelsius} boundary on measured inputs combined with held-input training reduced the PEACE$\to$Brauns error to about \SI{1.39}{\degreeCelsius}, which is still larger than the matched RL-only value of about \SI{1.20}{\degreeCelsius} under the same setting. Configurations are evaluated in both directions without direction-specific selection on the test data. Across the additional identification schemes (one-step and ridge identification, relative coordinates) and projection fractions, improvements over the main PR-RL configuration also remain specific to one transfer direction.

\section{Prediction horizons beyond training and selection}
\label{si:long}

This section reports the endpoint RMSE against the measured temperature at 60, 90, and \SI{120}{\minute}, to assess accuracy beyond the training and selection horizons. It uses a separate configuration in which training and validation-based selection use accuracy targets at horizons of up to \SI{30}{\minute}. The three evaluation horizons therefore exceed those used for training and selection; one Brauns condition differs from the main configuration, whose configuration selection additionally uses the validation RMSE at 60, 90, and \SI{120}{\minute} (the `Statistical analysis' section of Methods). The \SI{120}{\minute} residual gain of the main text concerns the difference between two predictions under the same inputs and is distinct from the \SI{120}{\minute} prediction error reported here. For this comparison, the matched RL-only model was trained under the same fixed configuration, with twenty new seeds per condition; the PEACE \SI{120}{\minute} results use the two sessions with complete evaluation windows.

Supplementary Table~\ref{tab:si-long} lists the six stored final comparisons.

\begin{table}[!htbp]
\def\baselinestretch{1}\normalsize\footnotesize
\caption{\textbf{Endpoint RMSE at horizons beyond training and selection.} Endpoint RMSE at horizons beyond training and selection: PR-RL versus the matched RL-only model, twenty seeds per condition, in \si{\degreeCelsius}. Differences are PR-RL minus the matched RL-only model with the stored Bonferroni-adjusted simultaneous confidence intervals of this six-comparison family (error probability 5\%); practical judgments use the \SI{0.10}{\degreeCelsius} margin, and statistical superiority or inferiority is judged against zero. Non-inferiority (NI) is recorded separately.}
\label{tab:si-long}
\centering
\footnotesize
\setlength{\tabcolsep}{4pt}
\resizebox{\textwidth}{!}{%
\begin{tabular}{@{}lccccccc@{}}
\toprule
Dataset & Horizon (\si{\minute}) & PR-RL & Matched RL-only & Difference & Simultaneous CI & Practical & Statistical; NI \\
\midrule
Brauns & 60 & 0.3711 & 0.3512 & $+0.0199$ & $[+0.0049,\,+0.0350]$ & Equivalent & Inferior; NI \\
Brauns & 90 & 0.3203 & 0.4165 & $-0.0962$ & $[-0.1377,\,-0.0548]$ & Undecided & Superior; NI \\
Brauns & 120 & 0.3631 & 0.5117 & $-0.1486$ & $[-0.2035,\,-0.0937]$ & Undecided & Superior; NI \\
PEACE & 60 & 5.7031 & 5.8505 & $-0.1474$ & $[-0.3384,\,+0.0436]$ & Undecided & Undecided; NI \\
PEACE & 90 & 5.9024 & 5.5757 & $+0.3268$ & $[+0.0687,\,+0.5848]$ & Undecided & Inferior; NI not confirmed \\
PEACE & 120 & 4.3784 & 4.1186 & $+0.2598$ & $[+0.0217,\,+0.4979]$ & Undecided & Inferior; NI not confirmed \\
\bottomrule
\end{tabular}}
\end{table}

Of the six comparisons, one is practically equivalent and five remain undecided at the \SI{0.10}{\degreeCelsius} margin. Statistically, PR-RL is superior at the Brauns 90 and \SI{120}{\minute} horizons, inferior at the Brauns \SI{60}{\minute} and the PEACE 90 and \SI{120}{\minute} horizons, and undecided at the PEACE \SI{60}{\minute} horizon. Accuracy beyond the training and selection horizons therefore depends on the dataset and the horizon. All horizons use the same engineering margin as the main text.

For reference, Supplementary Table~\ref{tab:si-long-ref} lists the earlier endpoint RMSE of the three configuration-selection seeds (11, 23, and 37); these values are not merged with the twenty-seed cohort.

\begin{table}[!htbp]
\def\baselinestretch{1}\normalsize\small
\caption{\textbf{Reference endpoint RMSE of the configuration-selection seeds.} Reference endpoint RMSE of the three configuration-selection seeds (11, 23, 37), in \si{\degreeCelsius}, averaged with equal weights over sessions and seeds. These earlier values are reported for reference only and are not merged with the twenty-seed cohort of Supplementary Table~\ref{tab:si-long}.}
\label{tab:si-long-ref}
\centering
\small
\begin{tabular}{@{}lccc@{}}
\toprule
Dataset & Horizon (\si{\minute}) & PR-RL & Matched RL-only \\
\midrule
Brauns & 60 & 0.3745 & 0.3323 \\
Brauns & 90 & 0.3179 & 0.4524 \\
Brauns & 120 & 0.3603 & 0.5845 \\
PEACE & 60 & 5.7514 & 5.9662 \\
PEACE & 90 & 5.9765 & 5.6263 \\
PEACE & 120 & 4.4464 & 4.3141 \\
\bottomrule
\end{tabular}
\end{table}

\end{document}